\documentclass[letterpaper]{article} 
\usepackage{aaai25}  
\usepackage{times}  
\usepackage{helvet}  
\usepackage{courier}  
\usepackage[hyphens]{url}  
\usepackage{graphicx} 
\usepackage{natbib}  
\usepackage{caption} 
\usepackage{algorithm}
\usepackage{algorithmic}
\usepackage{graphicx}

\usepackage{amsmath}
\usepackage{algorithm}
\usepackage{algorithmic}
\usepackage[utf8]{inputenc} 
\usepackage{graphicx}
\usepackage[T1]{fontenc}    
\usepackage{url}            
\usepackage{booktabs}       
\usepackage{amsfonts}       
\usepackage{nicefrac}       
\usepackage{microtype}      
\usepackage{xcolor}         
\usepackage{subcaption}
\usepackage{multirow}
\usepackage{multicol}
\usepackage{enumitem}
\usepackage{xcolor} 
\usepackage{color, colortbl}
\usepackage{subfiles}
\definecolor{mygray}{gray}{.9}

\newcommand{\answerYes}[1]{\textcolor{blue}{[Yes]}}
\newcommand{\answerPartial}[1]{\textcolor{blue}{[Partial]}}
\newcommand{\answerNo}[1]{\textcolor{red}{[No]}}
\newcommand{\answerNA}[1]{\textcolor{gray}{[NA]}}
\usepackage{newfloat}
\usepackage{listings}
\DeclareCaptionStyle{ruled}{labelfont=normalfont,labelsep=colon,strut=off} 
\floatstyle{ruled}
\newfloat{listing}{tb}{lst}{}
\floatname{listing}{Listing}
\title{ACCon: Angle-Compensated Contrastive Regularizer for  Deep Regression}
\author {
    Botao Zhao$^{*}$  ,
    Xiaoyang Qu\thanks{These authors contributed equally to this work.}, 
    Zuheng Kang,
    Junqing Peng,
    Jing Xiao,
    Jianzong Wang\thanks{Corresponding author}
}
\affiliations {
    Ping An Technology (Shenzhen) Co., Ltd. \\
    jzwang@188.com, zhao\_bot@163.com,\\
    \{quxiaoyang343, kangzuheng896, pengjq, xiaojing661\}@pingan.com.cn
}

\begin{document}

\maketitle

\begin{abstract}

	In deep regression, capturing the relationship among continuous labels in feature space is a fundamental challenge that has attracted increasing interest.
	Addressing this issue can prevent models from converging to suboptimal solutions across various regression tasks, leading to improved performance, especially for imbalanced regression and under limited sample sizes.
	However, existing approaches often rely on order-aware representation learning or distance-based weighting.
	In this paper, we hypothesize a linear negative correlation between label distances and representation similarities in regression tasks.
	To implement this, we propose an angle-compensated contrastive regularizer for deep regression, which adjusts the cosine distance between anchor and negative samples within the contrastive learning framework.
	Our method offers a plug-and-play compatible solution that extends most existing contrastive learning methods for regression tasks.
	Extensive experiments and theoretical analysis demonstrate that our proposed angle-compensated contrastive regularizer not only achieves competitive regression performance but also excels in data efficiency and effectiveness on imbalanced datasets.

\end{abstract}
%
\begin{figure*}
	\centering
	\label{fig0}
	\includegraphics[width=0.93\linewidth]{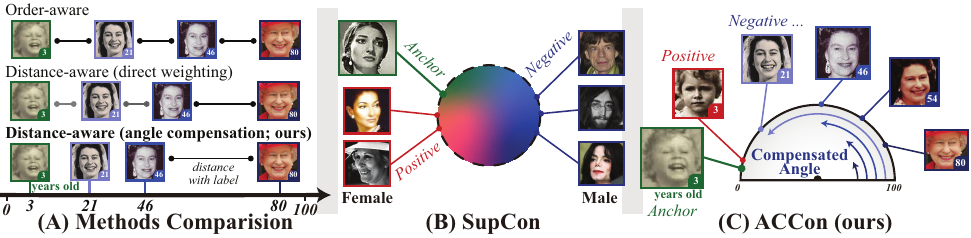}
	\caption{
		The subfigure (A) illustrates two types of representation learning for regression: order-aware and distance-aware. The subfigure (B) depicts the supervised contrastive learning for gender classification, where all samples from the same class are treated as positives and contrasted against negatives. The subfigure (C) presents the proposed angle-compensated supervised contrastive learning approach for age regression, which projects the input onto a semi-hypersphere while preserving the label relationship information.}
	\label{fig1}
	\vspace{-0.5cm}
\end{figure*}

\section{Introduction}

Regression, a fundamental machine learning task, is employed when the prediction target is continuous.
In the last ten years, deep regression has emerged as a more effective approach than traditional regression methods across various domains, including computer vision \cite{wang2022improving}, natural language process \cite{chandrasekaran2021evolution}.
Recent studies in deep regression have predominantly focused on developing network structures for specific application scenarios \cite{lee2021patch,chen2021time,tabelini2021polylanenet}.
However, these task-specific architectures are difficult to adapt to different tasks or domains, which remains a non-trivial research question.

A in-depth examination of deep regression by \cite{lathuiliere2019comprehensive} revealed that a well-tuned general-purpose network can achieve results close to the state-of-the-art (SOTA) models, potentially obviating the need for more intricate and specialized regression models.
In a standard deep regression learning (SDL) task, the representation is first extracted by the backbone network and subsequently processed by passing through a fully connected layer to generate the prediction for the target.
The mean squared error (MSE) loss \cite{lathuiliere2019comprehensive} is broadly acknowledged as the most frequently utilized loss function in deep regression.
Additionally, researchers have explored alternative loss functions, including L1 loss, Huber loss \cite{huber1992robust}, Tukey loss \cite{belagiannis2015robust}.
In an SDL, enhancing the quality of learned representations is a key approach to improving performance.

Given the continuous nature of regression targets, recent research in representation learning has increasingly focused on capturing the nuanced relationships within the label space.
\emph{These methods can be broadly categorized into: order-aware~\cite{gong2022ranksim, zha2024rank}, and distance-aware approaches~\cite{keramati2023conr, dai2021adaptive}.}
Order-aware methods typically rely on ranking-based techniques to constrain models to obtain order-aware representations.
However, these methods have a major flaw (Fig. \ref{fig1} (A), first row), as regression targets involve not only order information but also distance information (Fig. \ref{fig1} (A), second row).
For example, consider face images with ages 2, 21, 46, and 80; ranking-based methods would generally assign these labels' representations to a scale with equal intervals, which might incorrectly reflect the true relationships among them.
Distance-aware methods, on the other hand, aim to maintain the representation similarity among samples proportional to their corresponding label distances.
Current distance-aware approaches commonly apply weighted cosine similarity between the anchor and negative pairs within the contrastive learning framework \cite{keramati2023conr, dai2021adaptive}.
However, we posit that direct weighting may not be optimal due to the nonlinearity of the cosine function.
This nonlinearity can lead to unevenly distributed representations on the hypersphere, potentially distorting the relationship between labels and features.
\textit{In this paper, we introduce angle compensation as a refinement to direct weighting, aiming to advance distance-aware representation learning in regression tasks through a more nuanced and effective approach.}

To this end, we considered building upon supervised contrastive learning (SupCon), an approach that has seen rapid development with numerous notable contributions \cite{ermolov2021whitening, chen2020simple, jiang2020multi}.
As shown in Fig. \ref{fig1} (B), SupCon seeks to minimize the distance between an anchor and a "positive" sample (belonging to the same class) within the embedding space, while also maximizing the distance between the anchor and several "negative" samples (from different classes). However, the application of these methods has been mainly focused on classification and segmentation tasks \cite{khosla2020supervised,sun2023contrastive,seifi2024ood}.
Extending these approaches to regression tasks presents several challenges, a key issue being the effective handling of varying distances among negative pairs.
In regression tasks, for instance, negative samples with labels such as 46 and 80 are expected to exhibit different similarities to an anchor with label 3.
To address this gap, we propose an angle-compensated contrastive learning (ACCon) method to fill this gap and achieve effective distance-aware representation learning for regression. Our main contributions are as follows:

\begin{itemize}[leftmargin=0.5cm]
	\item We address the challenge of learning distance-aware representations for deep regression by proposing ACCon, a straightforward yet effective method.
	      ACCon can be integrated into various contrastive learning-based approaches to enhance performance.
	      Specifically, ACCon projects the input onto a semi-hypersphere, preserving the relationships among labels, as illustrated in Fig. \ref{fig1} (C).
	\item We provide theoretical proof that optimizing $\mathcal{L}_{\mathrm{ACCon}}$ results in features located on the semi-hypersphere according to the distances among labels.
	\item We perform comprehensive experiments across three datasets from both computer vision and natural language processing fields, showcasing enhanced performance relative to current state-of-the-art techniques.
\end{itemize}

\section{Related Work}
\subsection{Deep Regression}
Many researchers have focused on deep regressions over the past decades.
Some researchers formulate regression as a classification problem \cite{gao2017deep, shi2023deep, rogez2017lcr, pan2018mean}.
However, this formulation fails to fully leverage the relationship among labels and introduces a trade-off between optimization complexity and method performance. Recent advancements have proposed various methods to address imbalance issues in deep regression \cite{yang2021delving,ren2022balanced}.
Although some applications aim to enhance model performance in balanced settings, such as disease prediction, many certain real-world scenarios require models to perform effectively on imbalanced naturally sampled data, where long-tail samples may be considered outliers and potentially disregarded.
Several studies have proposed regularizers to constrain the embedding space, including approaches that model uncertainty within the embedding space \cite{li2021learning} and methods that learn higher-entropy feature spaces \cite{zhang2022improving}.
Recently, the concept of learning ordered features for regression has gained increased attention, with notable contributions such as Rank-N-Contrast \cite{zha2024rank} and RankSIM \cite{gong2022ranksim}.
\emph{In contrast to existing works, we propose a distance-aware representation learning method capable of capturing relationships among continuous labels in feature space.}

\subsection{Contrastive Learning}
Contrastive learning is a representation learning technique that projects features onto a hypersphere \cite{ermolov2021whitening,chen2020simple,jiang2020multi}.
The fundamental principle of contrastive learning involves attracting positive sample pairs while repelling negative ones.
In self-supervision, positive sample pairs are generally formed by different augmentations of the same sample, while negative pairs are made up of the anchor and other samples within the minibatch.
SupCon, an extension of contrastive learning in fully-supervised settings, has recently emerged as a promising approach, demonstrating significant advancements in image recognition and various classification tasks \cite{chen2020simple, sun2023contrastive, khosla2020supervised, liu2023multiple, seifi2024ood}.
In the SupCon framework, positive pairs are defined as samples from the same class, while negative pairs are samples from different classes.
However, the adaptation of contrastive learning to regression tasks presents several challenges, particularly in the definition of positive and negative pairs and the capture of relationships within the label space.
\cite{zha2024rank} proposed a redefinition of positive and negative pairs to achieve order-aware representation learning for regression, while \cite{dai2021adaptive} exploited a distance-weighted negative cosine similarity to achieve distance-aware representations.
\cite{keramati2023conr} introduced a novel approach that defines negative pairs as samples with dissimilar labels but similar predictions to the anchors. By combining this definition with distance-weighted negative cosine similarity, their method demonstrated robust performance in imbalanced regression tasks.
\emph{In contrast to these existing approaches, we refine distance-weighted negative cosine similarity by introducing angle compensation, striving to enhance distance-aware representation learning in regression tasks.}

\section{Methodology}
\subsection{Problem Setting}
In this paper, we consider the input $\left(\mathcal{X},\mathcal{Y}\right)=\left\{x_{i}, y_{i}\right\}_{i=1}^{N}$.
Similar to contrastive learning, our objective is to develop a feature representation network that could map the input $x_{i}$ to an L2-normalized d-dimensional embedding, $z_{i} \in \mathcal{S}^{d-1}$.
Furthermore, we assume that the learned representation $z \in Z$ should be distributed on the $\mathcal{S}^{d-1}$ such that each position corresponds to its label, as illustrated in Fig. \ref{fig1} (C).

Given the continuous nature of labels, we first partition the label space $\mathcal{Y}$ into $M$ bins with equal intervals, denoted as $\left[ y_{0}, y_{1} \right), \left[ y_{1}, y_{2}\right), ... ,\left[y_{M - 1}, y_{M}\right)$.
These bins represent the precision of the labels, which can be adjusted according to practical requirements.

We categorize samples from the same bin as positive pairs, while those from different bins are considered negative pairs. 
Specifically, for an anchor $x_{i}$, the set of the positive pairs is denoted as $\mathcal{P}(i):= \left \{ j \in \mathcal{B} | y_{i}=y_{j},j \neq i \right \}$, and the negative pairs are defined as $\mathcal{N}(i):= \left \{ j \in \mathcal{B} | y_{i} \neq y_{j} \right \}$.
Subsequently, we can obtain a naïve extension of supervised contrastive learning loss function \cite{ dai2021adaptive, keramati2023conr, khosla2020supervised} for regression tasks as follows:
\begin{equation}
	\label{eq:scl2}
	\mathcal{L}_{i} = -\frac{1}{|\mathcal{P}\left(i\right)|}\sum_{p \in \mathcal{P}\left(i\right)}   \mathrm{log} \frac{\mathrm{exp}\left(\cos\left(\theta_{i,p}\right) / \tau\right)}{\sum_{k \in \mathcal{N}\left(i\right) \cup \mathcal{P}\left(i\right)} \mathrm{exp}\left(\cos\left(\theta_{i,k}\right)/\tau\right)},
\end{equation}
where, $\theta_{i,k}$ denotes the angle between embedding $z_{i}$ and $z_{k}$, and $\cos\left(\theta_{i,k}\right) = z_{i}z_{k}^{T}$.
However, this formulation is unsuitable for regression tasks as it fails to capture the inherent relationships among labels.
As illustrated in Fig. \ref{fig1} (C), both samples with labels 46 and 80 are treated equivalently as negatives for an anchor with label 3. Yet, the dissimilarity between the anchor 3 and label 80 is significantly greater than that between the anchor and label 46.
Furthermore, due to the inherent monotonic nature of labels in regression tasks, mapping $x_{i}$ to a complete hypersphere faces great challenges.
Therefore, our objective is to \textit{develop a model that maps the learned representation $z$ distributed onto a semi-hypersphere, positioning them in accordance with their respective labels.}

\subsection{Angle-Compensated Supervised Contrastive Loss}

We propose an angle-compensated supervised contrastive loss to achieve our goal of preserving label relationships in feature space.
Firstly, we hypothesize that a linear negative correlation exists between label distances and representation similarities, implying that samples' representations should be positioned in alignment with their respective labels.
Based on this hypothesis, if we use cosine distance to measure representation similarities, we can derive the ideal angle, $\hat{\theta}$ between anchor and negatives as follows:
\begin{equation}
	\hat{\theta} = \frac{y_{\mathrm{neg}} - y_{\mathrm{anc}}}{\max\left(\mathcal{Y}\right)-\min\left(\mathcal{Y}\right)} \pi,
	\label{angle}
\end{equation}
where ,$y_{\mathrm{anc}}, y_{\mathrm{neg}}$ denote the label of anchor and negative sample, respectively.
For instance, given an anchor with a label of 21 and an age range from 0 to 100, the angles between the anchor and negative samples (3, 26, 54, 80) would be $-5.4^{\circ}$, $32.4^{\circ}$, $41.4^{\circ}$, $91.8^{\circ}$ and $138.6^{\circ}$, respectively.

Several studies have demonstrated that the representations of each class spontaneously collapse to the vertices of a regular simplex when the standard supervised contrastive loss reaches its minimum \cite{graf2021dissecting,zhu2022balanced}.
The SupCon loss effectively constrains the representations of anchors and negatives to be as far apart as possible within a minibatch, which is equivalent to constraining the included angle $\tilde{\theta}=\pi$.
This observation forms the basis for the formulation of Eq. \ref{angle}.
\begin{equation}
	\cos\left(\tilde{\theta}\right) = \cos\left(\hat{\theta} + \pi - \frac{y_{\mathrm{neg}} - y_{\mathrm{anc}}}{\max\left(\mathcal{Y}\right)-\min\left(\mathcal{Y}\right)}\pi\right).
	\label{eq4}
\end{equation}
Based on the Eq. \ref{eq:scl2}, we will get the angle-compensated supervised contrastive loss for sample $i$ as follows:
\begin{equation}
	\scriptsize
	\mathcal{L} _{i}^{\mathrm{ac}}=\frac{-1}{|\mathcal{P} \left( i \right) |}\sum_{p\in \mathcal{P} \left( i \right)}{\log}\frac{\exp \left( z_iz_{p}^{T}/\tau \right)}{\left( \begin{array}{c}
				\sum_{k\in \mathcal{P} \left( i \right)}{\exp}\left( z_iz_{k}^{T}/\tau \right)                                \\
				+\sum_{m\in \mathcal{N} \left( i \right)}{\exp}\left( \cos \left( \tilde{\theta} _{i,m} \right) /\tau \right) \\
			\end{array} \right)},
	\label{eq5}
\end{equation}
where, $\mathcal{P}(i)$ denotes the set of positive pairs for anchor $i$, and $\mathcal{N}(i)$ represents the negative set.
This formulation is equal to the SupCon loss, with the key distinction that it replaces the negative similarity with an angle-compensated version.
Considering a minibatch, the optimizition of Eq. \ref{eq5} aims to drive $\tilde{\theta}_{i,m} \rightarrow \pi$ and make the $\hat{\theta}_{i,m} \rightarrow \frac{y_{\mathrm{neg}} - y_{\mathrm{anc}}}{\max\left(\mathcal{Y}\right)-\min \left(\mathcal{Y}\right)} \pi$, aligning with our intended goal.
Then, the angle-compensated supervised contrastive loss $\mathcal{L}_{\mathrm{ACCon}}$ is defined as the mean value of $\left\{ \mathcal{L}^{\mathrm{ac}}_{i} \right\}_{i=1}^{2N}$:
\begin{equation}
	\mathcal{L}_{\mathrm{ACCon}} = \frac{1}{2N} \sum_{i}^{2N}  \mathcal{L}^{\mathrm{ac}}_{i}
	\label{sup1}
\end{equation}
where $N$ is the batch size.

To compute the $\mathcal{L}^{\mathrm{ac}}_{i}$, we should determine $\cos \left(\tilde{\theta}_{i,m}\right)$.
We firstly define the compensation angle $\varphi$ based on the Eq. \ref{eq4}:
\begin{equation}
	\varphi = \pi\left(1 -\frac{y_{\mathrm{neg}} - y_{\mathrm{anc}}}{\max\left(\mathcal{Y}\right)-\min\left(\mathcal{Y}\right)}\right).
	\label{eq6}
\end{equation}
Subsequently, we derive $\cos \left(\tilde{\theta}_{i,m}\right)$ as follows:
\begin{equation}
	\cos\left(\tilde{\theta}_{i,m}\right) = z_{i}z_{m}^{T}\cos\left(\varphi\right)-\left|\sin\left(\varphi\right)\right|\sqrt{1-\left(z_{i}z_{m}^{T}\right)^2 + \epsilon},
	\label{eq7}
\end{equation}
where $\epsilon$ is the smoothing term introduced to prevent gradient explosion.
A detail oderivation of $\cos\left(\tilde{\theta}_{i,m}\right)$ is provided in Appendix A1.

\subsection{Theoretical Insights}

In this section, we theoretically analyzed the efficiency of $\mathcal{L}_{\mathrm{ACCon}}$.
Firstly, we have derived a lower bound of $\mathcal{L}_{\mathrm{ACCon}}$ in Appendix A2.
\newtheorem{theorem}{Theorem}

\begin{theorem}[Lower bound of $\mathcal{L}_{\mathrm{ACCon}}$].
	$\mathrm{L}^{*} := \frac{1}{4N^{2}} \sum_{i=0}^{2N} \sum\limits_{m \in \mathcal{N}(i)} \cos\left(\tilde{\theta}_{i,m}\right)/\tau - \frac{\mathrm{log N / \tau}}{2N}$ is a lower bound of $\mathcal{L}_{\mathrm{ACCon}}$, i.e, $\mathcal{L}_{\mathrm{ACCon}} > \mathrm{L}^{*}$.
	\label{theorem1}
\end{theorem}
Minimizing the $\mathcal{L}_{\mathrm{ACCon}}$, will lead to the minimization of $\mathrm{L}^{*}$.
Optimizing $\mathrm{L}^{*}$ results in the convergence of $\hat{\theta}$ to $\frac{y_{\mathrm{neg}} - y_{\mathrm{anc}}}{\max\left(\mathcal{Y}\right)-\min\left(\mathcal{Y}\right)} \pi$, which aligns with our initial hypothesis.

\subsection{Deep Regression Framework}
\begin{figure}
	\centering
	\centering
	\begin{minipage}[b]{1\linewidth}
		\centering
		\centerline{\includegraphics[width=0.94\linewidth]{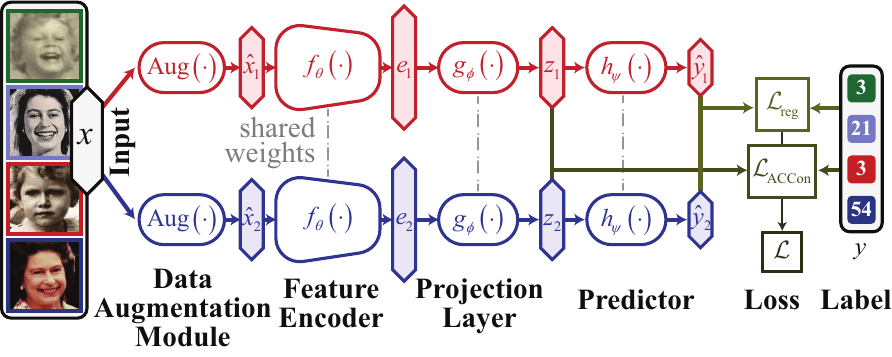}}
	\end{minipage}
	\caption{The frameworks of the deep regression with angle-compensated supervised contrastive.}
	\label{fig2}
	\vspace{-0.5cm}
\end{figure}

The proposed angle-compensated supervised contrastive loss integrates seamlessly into traditional supervised contrastive learning frameworks \cite{khosla2020supervised}, as demonstrated in Fig. \ref{fig2}, which depicts the two-view condition.
Our method is inherently scalable, allowing for the adjustment of data augmentations according to the specific requirements of the learning task.
In detail, the deep regression framework consists of four core components:
\begin{itemize}[leftmargin=0.5cm]
	\item Data augmentation module, $\mathrm{Aug}\left(\cdot\right)$: This augments the input $x$ to $\hat{x}_{1}$ and $\hat{x}_{2}$, randomly, with each augmentation providing a distinct view of the input data.
	\item Feature encoder, $f_{\theta}\left(\cdot\right)$: This component maps the augmented input to an embedding, denoted as $z=f_{\theta}\left(\hat{x}\right)$.
	      The architectural design of the encoder may vary depending on the specific application scenario.
	\item Projection layer, $g_{\phi}\left(\cdot\right)$: This layer projects the embedding $e$ to a representation vector $\hat{z}=g_{\phi}\left(z\right) \in \mathbb{R}^{d_{l}}$.
	      Typically, $g_{\phi}\left(\cdot\right)$ is implemented by a single linear layer, followed by $L_{2}$ normalization to ensure $\hat{z} \in \mathcal{S}^{d_{l}-1}$.
	\item Predictor, $ h_{\psi}\left(\cdot\right)$: The predictor maps the representation $\hat{z}$ to the label $\hat{y}$ utilizing a single linear layer without bias in our configuration.
\end{itemize}
After applying these modules, we obtain the extracted embedding $\hat{z}$ and the predicted target $\hat{y}$.
Subsequently, our model is trained through the simultaneous optimization of the regression loss function $\mathcal{L}_{\mathrm{reg}}$ and the angle-compensated supervised contrastive loss $\mathcal{L}_{\mathrm{ACCon}}$.
The final loss function is formulated as:
\begin{equation}
	\mathcal{L} = \mathcal{L}_{\mathrm{reg}} + \gamma \cdot \mathcal{L}_{\mathrm{ACCon}},
	\label{sumloss}
\end{equation}
where $\gamma$ denotes the weight coefficient.
In this study, we employ the mean square error (MSE) as $\mathcal{L}_{\mathrm{reg}}$ for the sentence similarity prediction task and the mean absolute error (MAE) as $\mathcal{L}_{\mathrm{reg}}$ for the age estimation task.
The comprehensive training process is delineated in Alg. \ref{alg:algorithm}.

\begin{algorithm}[t]
	\caption{Training of the deep regression with ACCon.}
	\label{alg:algorithm}
    \footnotesize

	\textbf{Input}: The minibatch data $ \left( X, Y \right) \subset \left( \mathcal{X}, \mathcal{Y} \right)$, Feature encoder $f_{\theta}$, Projection layer $g_{\phi}$, Predictor $h_{\psi}$. \\
	\textbf{Parameter}: The temperature index, $\tau$; Weight coefficient, $\gamma$; Max training epoch, $T$; Smoothing term, $\epsilon$.\\
	\textbf{Output}: $\theta^{*}, \phi^{*}, \psi^{*}$
	\begin{algorithmic}[1]  
		\STATE Let $t=0$.
		\STATE Randomly initialize $\theta_{0}, \phi_{0}, \psi_{0}$ \\
		\WHILE{$t < T$}
		\STATE draw two augmentation functions $\mathrm{Aug}_{1}\left(\cdot\right)$, $\mathrm{Aug}_{2}\left(\cdot\right)$\\
		\COMMENT{\textcolor{gray} {Forwad process}} \\

		\STATE $\Hat{X}^{c}\leftarrow \mathrm{cat}\left([\mathrm{Aug}_{1}\left(X\right),\mathrm{Aug}_{2}\left(X\right)], \mathrm{dim}=0\right)$ \\
		\STATE $E \leftarrow f_{\theta}\left(\Hat{X}^{c}\right)$ \\
		\STATE $Z \leftarrow g_{\phi}\left(E\right)$ \\
		\STATE $\tilde{Z} \leftarrow Z/\left\|Z\right\|$ \\

		\COMMENT{\textcolor{gray}{Calculation of loss function}} \\
		\STATE $\tilde{Y} \leftarrow \mathrm{cat}\left([\hat{Y}, \hat{Y}], \mathrm{dim}=0\right)$ \\
		\STATE Calculate $\varphi$ by  Eq. \ref{eq6} \\
		\STATE Calculate $\cos\left(\tilde{\theta}\right)$ based on Eq. \ref{eq7} \\
		\STATE Calculate $\mathcal{L}_{\mathrm{ACCon}}$ based on Eq. \ref{eq5} and Eq. \ref{sup1} \\
		\STATE Calculate the $\mathcal{L}_{\mathrm{reg}}$ between $h_{\psi}\left(Z\right)$ and $Y$\\
		\STATE Calculate loss function $\mathcal{L}$ by Eq. \ref{sumloss}   \\

		\COMMENT{\textcolor{gray}{Optimization}}
		\STATE Calculate $\theta_{t+1}$, $\phi_{t+1}$, $\psi_{t+1}$ by gradient descent of $\mathcal{L}$ \\
		\STATE $t=t+1$ \\
		\ENDWHILE
		\STATE $\theta^{*},\phi^{*},\psi^{*} \leftarrow \theta_{T}, \phi_{T}, \psi_{T}$
		\STATE \textbf{return $\theta^{*}$, $\phi^{*}$, $\psi^{*}$}
	\end{algorithmic}
\end{algorithm}

\begin{table*}[t]
	\setlength{\tabcolsep}{3pt}
	\renewcommand{\arraystretch}{1.0}

	\centering
	\resizebox{\linewidth}{!}{
		\begin{tabular}{lccccccccc}
			\toprule   
			\multicolumn{1}{l}{\multirow{2}{*}{Methods}} & \multicolumn{3}{c}{\multirow{1}{*}{AgeDB-Natural}} & \multicolumn{3}{c}{\multirow{1}{*}{STS-B-Natural}} & \multicolumn{3}{c}{\multirow{1}{*}{IMDB-WIKI-Natural}}                                                                                                                                                                                                                               \\

			\cmidrule(lr){2-4} \cmidrule(lr){5-7}\cmidrule(lr){8-10}
			\multicolumn{1}{c}{}                         & MAE  $\downarrow$                                  & G-means $\downarrow$                               & $\mathrm{R}^{2}$ $\uparrow$                            & MAE  $\downarrow$                  & G-means $\downarrow$               & Pearson $\uparrow$
			                                             & MAE  $\downarrow$                                  & G-means $\downarrow$                               & $\mathrm{R}^{2}$ $\uparrow$                                                                                                                                                                                                                                                          \\
			\midrule   
			Vanilla                                      & 7.044                                              & 4.449                                              & 0.715                                                  & 0.798                              & 0.516                              & 0.756                              & 5.642                              & 3.251                              & 0.646                              \\
			NaïveSupCon \cite{khosla2020supervised}      & 7.070                                              & 4.549                                              & 0.718                                                  & 0.788                              & 0.518                              & 0.764                              & 5.542                              & 3.221                              & 0.642                              \\
			AdaSupCon \cite{dai2021adaptive}             & 6.973                                              & 4.438                                              & 0.725                                                  & 0.775                              & 0.503                              & 0.763                              & 5.535                              & 3.210                              & 0.652                              \\
			BMSE \cite{ren2022balanced}                  & 6.999                                              & 4.384                                              & 0.721                                                  & /                                  & /                                  & /                                  & 5.589                              & 3.235                              & 0.638                              \\
			RankSim \cite{gong2022ranksim}               & 6.914                                              & 4.404                                              & 0.724                                                  & 0.786                              & 0.519                              & 0.756                              & 5.481                              & 3.198                              & 0.654                              \\
			RNC \cite{zha2024rank}                       & 6.793                                              & 4.298                                              & 0.732                                                  & 0.782                              & 0.512                              & 0.762                              & 5.483                              & 3.182                              & 0.656                              \\
			ConR \cite{keramati2023conr}                 & 6.808                                              & 4.324                                              & 0.733                                                  & 0.795                              & 0.517                              & 0.760                              & 5.492                              & 3.199                              & 0.655                              \\
			\textbf{ACCon (Ours)}                        & \textbf{6.724}                                     & \textbf{4.245}                                     & \textbf{0.741}                                         & \textbf{0.724}                     & \textbf{0.469}                     & \textbf{0.791}                     & \textbf{5.432}                     & \textbf{3.157}                     & \textbf{0.668}                     \\
			\midrule   
			Ours vs. Vanilla                             & \textcolor{teal}{$\uparrow$4.54\%}                 & \textcolor{teal}{$\uparrow$4.59\%}                 & \textcolor{teal}{$\uparrow$3.64\%}                     & \textcolor{teal}{$\uparrow$9.27\%} & \textcolor{teal}{$\uparrow$9.11\%} & \textcolor{teal}{$\uparrow$4.63\%} & \textcolor{teal}{$\uparrow$3.72\%} & \textcolor{teal}{$\uparrow$2.89\%} & \textcolor{teal}{$\uparrow$3.41\%} \\
			Ours vs. SOTA                                & \textcolor{teal}{$\uparrow$1.02\%}                 & \textcolor{teal}{$\uparrow$1.23\%}                 & \textcolor{teal}{$\uparrow$1.08\%}                     & \textcolor{teal}{$\uparrow$6.58\%} & \textcolor{teal}{$\uparrow$6.76\%} & \textcolor{teal}{$\uparrow$3.53\%} & \textcolor{teal}{$\uparrow$1.91\%} & \textcolor{teal}{$\uparrow$1.10\%} & \textcolor{teal}{$\uparrow$1.83\%} \\
			\bottomrule   
		\end{tabular}}
	\label{tab11}
    	\caption{Performance comparison with other regularizers for regression on the AgeDB-Natural, STS-Natural, and IMDB-WIKI-Natural. We trained each baseline model and the detailed information as described in Appendix B.}
	\vspace{-1.1em}
\end{table*}

\section{Experiments and Results}

\subsection{Experiments Setup}
\label{ExperimentsSetup}
\subsubsection{Datasets:}
To rigorously evaluate our method, we selected three diverse datasets:
(1) \textbf{AgeDB} \cite{moschoglou2017agedb}: An age estimation dataset comprising 16,488 facial images;
(2) \textbf{IMDB-WIKI} \cite{rothe2018deep}: A large-scale facial age dataset containing 523,000 images with corresponding age labels;
(3) \textbf{STS-B} \cite{cer2017semeval,wang2018glue}: A natural language dataset consisting of 7,249 sentence pairs, extracted from the Semantic Textual Similarity (STS) Benchmark.
To ensure a comprehensive assessment of our proposed method, we employed 2 distinct sampling strategies for partitioning the datasets into training, validation, and test sets:

\noindent
(1) \emph{Balanced Sampling}: Following the benchmark established by \cite{yang2021delving}, we partitioned the AgeDB dataset into 12,208 training samples, 2,140 validation samples, and 2,140 test samples.
For IMDB-WIKI, we allocated 191,500 images for training and 11,000 images each for validation and testing.
From STS-B, we sampled 1,000 pairs each for validation and testing. This benchmark ensures a balanced label distribution in the validation and test sets, as illustrated in Appendix Figure 1.
Similar to Yang et al.'s benchmark, we denote these balanced datasets as AgeDB-DIR, IMDB-WIKI-DIR, and STS-B-DIR.

\noindent
(2) \emph{Natural Sampling}: We used the same dataset splitting ratio but randomized the division of training, validation, and test datasets, thereby preserving similar label distributions across all three sets (Appendix Figure 1).
We denote the datasets obtained through natural sampling as AgeDB-Natural, IMDB-WIKI-Natural and STS-B-Natural.

\begin{figure*}[t]
	\centering
	\begin{minipage}[t]{0.47\textwidth}
		\centering
		\centerline{\includegraphics[width=8.3cm]{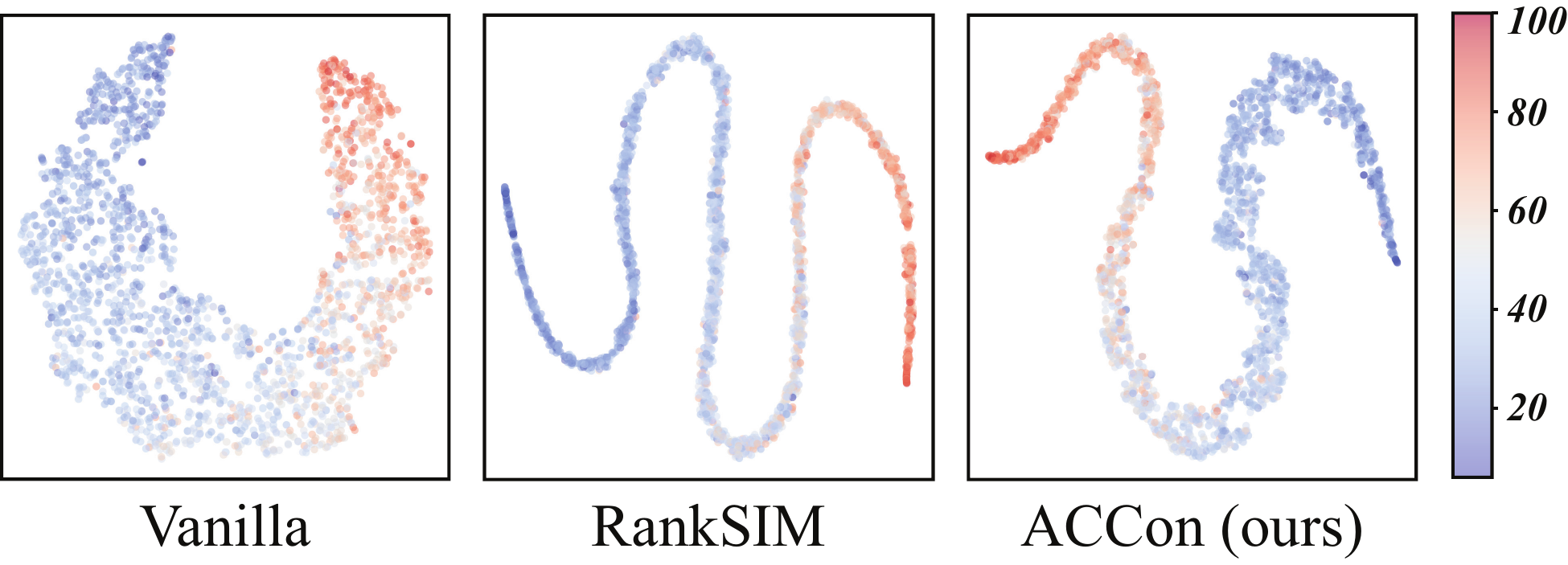}}
		\centerline{(a) t-SNE visualization} \medskip
	\end{minipage}
	\begin{minipage}[t]{0.52\textwidth}
		\centering
		\centerline{\includegraphics[width=9.2cm]{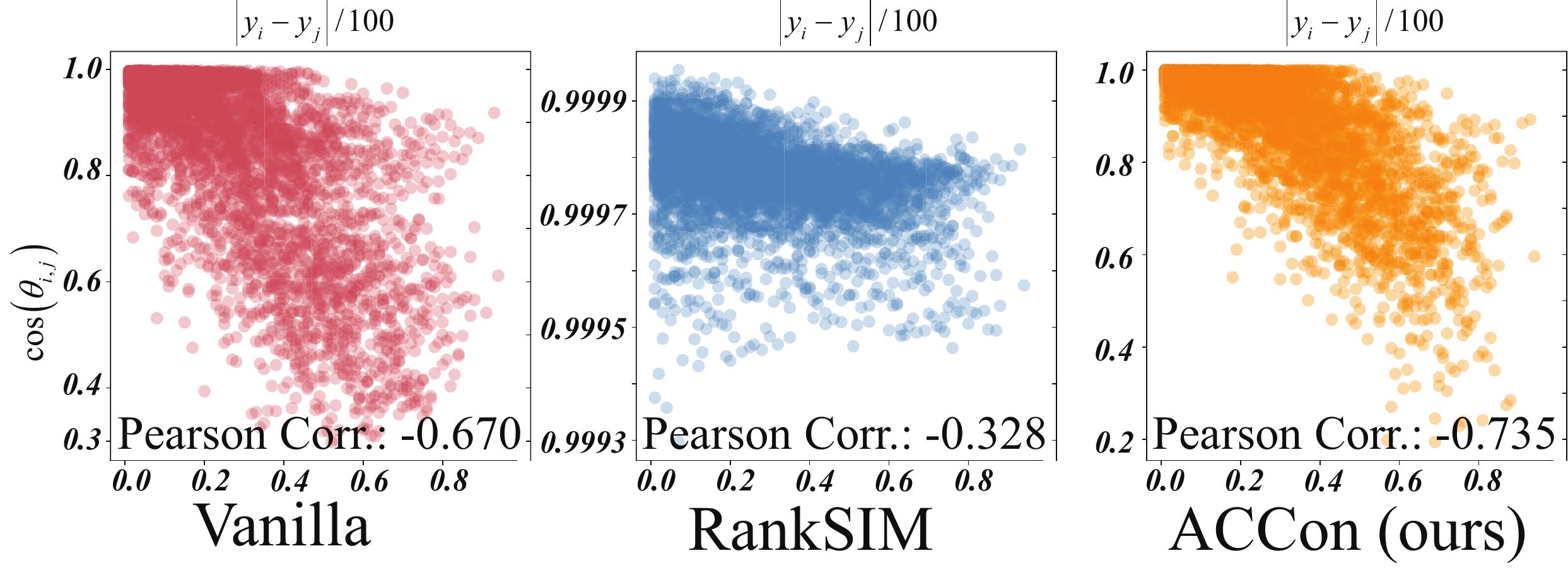}}
		\centerline{(b) Distributions of $\cos\left(\theta_{i,j}\right)$ and $\left|y_{i}-y_{j}\right|/100$}
	\end{minipage}
	\vspace{-0.3cm}

	\caption{The visualization and quantitation analysis of feature representations.
	The subfigure (a) is the t-SNE visualization of feature space on the AgeDB-natural test dataset. The subfigure (b) depicts the joint distribution of $\cos\left(\theta_{i,j}\right)$ and the distance of labels $\left|y_{i}-y_{j}\right|/100$ on the AgeDB-natural test dataset using kernel density estimation.}
	\label{fig4}
	\vspace{-0.3cm}
\end{figure*}

\begin{figure}[h]
	\centering
	\centering
	{\includegraphics[width=7.5cm]{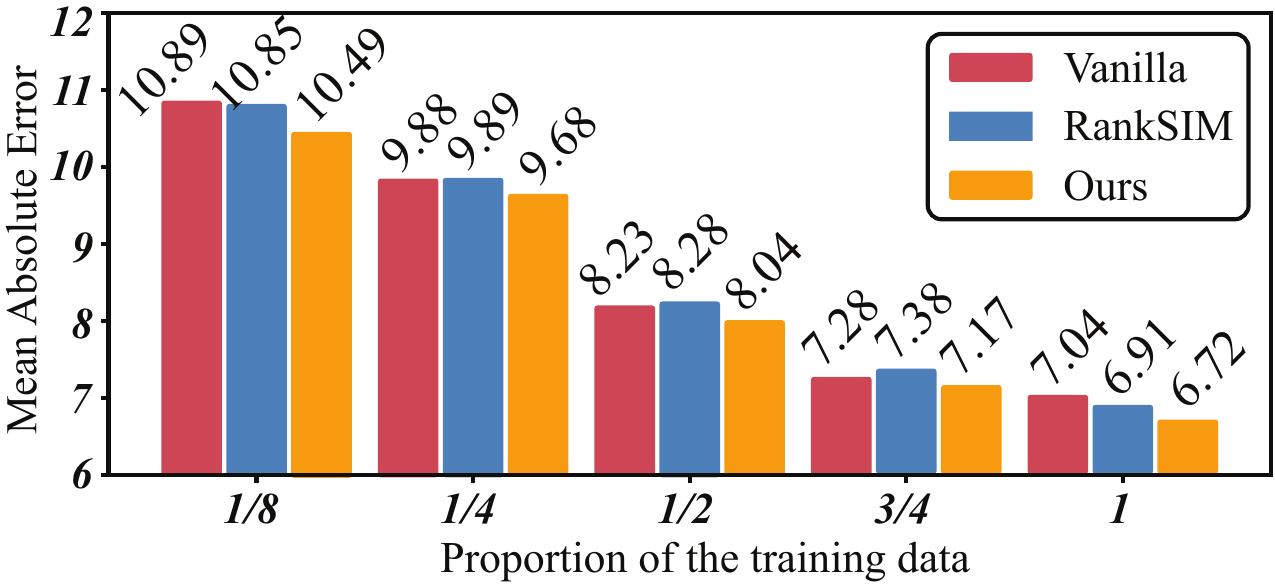}}
	\caption{The performance comparison on AgeDB-Natural, when reducing the training dataset.}
	\label{data_efficency}
	\vspace{-0.6cm}
\end{figure}

\subsubsection{Metrics:}

For quantitative evaluation, we utilized a diverse set of metrics including Mean Absolute Error (MAE), Geometric Mean (GM) defined as $\left(\prod \limits_{i=1}^n e_{i}\right)^{\frac{1}{n}}$ where $e_{i}$ represents the $L_{1}$ error for the $i^{th}$ sample, coefficient of determination $\mathrm{R}^{2}$ ($1-\frac{\mathrm{MSE}\left(y, \hat{y}\right)}{\mathrm{VAR}\left(y\right)}$), indicating the proportion of explainable variance, Mean Square Error (MSE), and Pearson correlation. The division of test set into many-shot, medium-shot, and few-shot categories follows the same protocol as outlined in \cite{yang2021delving}.

\subsection{Baselines}
We compared our method with state-of-the-art (SOTA) approaches under both natural and balanced sampling conditions. To guarantee a fair evaluation, we utilized standard network architectures consistent with those outlined by \cite{yang2021delving}.
Specifically, ResNet-50 was utilized as the backbone for age estimation tasks, while BiLSTM+GloVe word embeddings were employed for sentence similarity prediction.

For STS-B-Natural, IMDB-WIKI-Natural, and AgeDB-Natural datasets, our method is compared with,
\textbf{Vanilla}: using MAE or MSE loss function;
\textbf{Naïve SupCon}: adopting SupCon for deep regression, as in Eq. \ref{eq:scl2};
\textbf{Adaptive SupCon (AdaSupCon)}: proposing adaptive-margin contrastive loss for image regression \cite{dai2021adaptive};
\textbf{RankSIM}: exploiting ranking similarity regularization \cite{gong2022ranksim};
\textbf{BMSE}: modifying MSE loss for deep imbalanced regression \cite{ren2022balanced};
\textbf{RNC}: an order-aware method \cite{zha2024rank};
\textbf{ConR}: exploiting contrastive learning for imbalanced regression \cite{keramati2023conr}.

For STS-B-DIR, IMDB-WIKI-DIR, and AgeDB-DIR datasets, our approach was benchmarked against Vanilla, NaïveSupCon, AdaSupCon, RankSIM, and ConR.
Additionally, comparisons were made with re-weighting methods such as Label Distribution Smooth (LDS) and Focal-R \cite{yang2021delving}, integrated with the baseline described above.
Typically, Re-weighting adjusts the loss function to address label imbalance.
So, we tested inverse-frequency weighting (\textbf{INV}) and its square-root variant (\textbf{SQINV}).
LDS employs re-weighting based on smoothed label distribution, while Focal-R utilizes a continuous function mapping absolute error to the scaling factor.
Detailed experimental setup, including training protocols, dataset specifications, and hyperparameters, are provided in Appendix B.

\subsection{Basic Performance}

We implement the vanilla model and five advanced regularizer methods tailored for regression tasks: NaïveSupCon \cite{khosla2020supervised}, AdaSupCon \cite{dai2021adaptive}, BMSE \cite{ren2022balanced}, RankSim \cite{gong2022ranksim}, RNC \cite{zha2024rank} and ConR \cite{keramati2023conr}.
All models were re-trained by us, with training details described in Appendix B.
Notably, BMSE proved challenging to train on STS-B-Natural; consequently, its evaluation on this dataset was omitted.
As evidenced in Table \ref{tab11}, our method outperformed all compared methods on AgeDB-Natural, STS-Natural, and IMDB-WIKI-Natural datasets.
Compared to the Vanilla model, our approach achieved MAE improvements of 4.54\%, 9.27\%, and 3.72\%, respectively.
Furthermore, our method attained SOTA performance across all three datasets, with MAE improvements of 1.02\%, 6.58\%, and 1.91\%, respectively.

\subsection{Performance for Imbalanced Regression}

We evaluated the efficacy of our proposed method on imbalanced regression tasks using the AgeDB-DIR, STS-B-DIR, and IMDB-WIKI-DIR datasets.
Furthermore, we plugged our approach with BMSE, SqrtINV, Focal-R, and ConR.
As shown in Table \ref{tab22}, our method demonstrated significant improvement over the vanilla model, indicating the effectiveness of ACCon.
Besides, comparative analysis with LDS, NaïveSupCon, AdaSupCon, ConR, and RankSIM reveals the competitiveness of our approach in addressing deep imbalanced regression.

Our method outperformed most baseline methods across the datasets, with the exception of STS-B-DIR.
We hypothesize that this exception may be attributed to the limited amount of data and constraints on data augmentation techniques for the semantic textual similarity task.
Additionally, we assessed the performance of our method when plugged with other deep imbalanced regression methods.
Results indicate that incorporation with ConR, SqrtINV, and Focal-R led to performance enhancements, particularly in the medium-shot and few-shot categories.

\begin{table*}[t]
	\setlength{\tabcolsep}{5pt}
	\renewcommand{\arraystretch}{0.95}

	\centering

	\label{Comparison1}
	\resizebox{\linewidth}{!}{
		\begin{tabular}{lcccccccccccc}
			\toprule   
			\multicolumn{1}{l}{\multirow{2}{*}{Methods}} & \multicolumn{4}{c}{\multirow{1}{*}{AgeDB-DIR} (MAE)} & \multicolumn{4}{c}{\multirow{1}{*}{IMDB-WIKI-DIR (MAE)}} & \multicolumn{4}{c}{\multirow{1}{*}{STS-B-DIR (MSE)}}                                                                                                                                                             \\
			\cmidrule(lr){2-5} \cmidrule(lr){6-9} \cmidrule(lr){10-13}
			\multicolumn{1}{c}{}                         & \multicolumn{1}{c}{All}                              & \multicolumn{1}{c}{Many}                                 & Media                                                & Few            & All            & Many           & Media          & Few            & All             & Many           & Media          & Few              \\

			\midrule   
			Vanilla                                      & 7.77                                                 & 6.62                                                     & 9.55                                                 & 13.67          & 8.06           & 7.23           & 15.12          & 26.33          & 0.974           & 0.851          & 1.520          & 0.984            \\

			+LDS                                         & 7.67                                                 & 6.98                                                     & 8.86                                                 & 10.89          & 7.83           & 7.30           & 12.43          & 22.51          & 0.914           & 0.819          & 1.319          & 0.955            \\

			+NaïveSupCon                                 & 7.51                                                 & 6.50                                                     & 9.23                                                 & 12.36          & 7.96           & 7.27           & 13.82          & 23.34          & 0.962           & 0.893          & 1.187          & 1.147            \\

			+AdaSupCon                                   & 7.29                                                 & 6.43                                                     & 8.59                                                 & 11.80          & 7.76           & 7.16           & 12.52          & 23.96          & 0.983           & 0.956          & 1.062          & 1.073            \\

			+RankSIM                                     & 7.13                                                 & 6.51                                                     & 8.17                                                 & 10.12          & 7.72           & 6.93           & 14.48          & 25.38          & 0.873           & 0.908          & 0.767          & 0.705            \\

			+ConR                                        & 7.20                                                 & 6.50                                                     & 8.04                                                 & 9.73           & 7.33           & 6.75           & 11.99          & 22.22          & 0.937           & 0.874          & 1.172          & 1.044            \\

			\textbf{+ACCon}                              & \textbf{6.81}                                        & {\textbf{6.26}}                                          & 7.56                                                 & 9.91           & 7.20           & \textbf{6.61}  & 11.92          & 22.54          & \textbf{0.802 } & \textbf{0.812} & 0.756          & 0.765            \\

			\textbf{+ACCon+ConR}                         & 6.82                                                 & 6.36                                                     & \textbf{7.52}                                        & \textbf{9.22 } & \textbf{7.19 } & 6.65           & \textbf{11.52} & \textbf{21.64} & 0.820           & 0.845          & \textbf{0.721} & \textbf{0.699 }  \\

			\midrule
			SqrtINV                                      & 7.81                                                 & 7.16                                                     & 8.80                                                 & 11.20
			                                             & 7.87                                                 & 7.24                                                     & 12.44                                                & 22.76          & 1.005          & 0.894          & 1.482          & 1.046                                                                                 \\

			+LDS                                         & 7.67                                                 & 6.98                                                     & 8.86                                                 & 10.89          & 7.83           & 7.31           & 12.43          & 22.51          & 0.914           & 0.819          & 1.319          & 0.955            \\

			+LDS+NaïveSupCon                             & 7.54                                                 & 6.94                                                     & 8.20                                                 & 11.44          & 7.81           & 7.26           & 12.23          & 22.25          & 1.030           & 0.998          & 1.133          & 1.125            \\

			+LDS+AdaSupCon                               & 7.43                                                 & 6.82                                                     & 8.24                                                 & 10.93          & 7.75           & 7.18           & 12.36          & 22.06          & 0.994           & 0.974          & 1.098          & 0.972            \\

			+LDS+RankSIM                                 & 6.99                                                 & 6.38                                                     & 7.88                                                 & 10.23          & 7.57           & 7.00           & 12.16          & 22.44          & 0.889           & 0.911          & 0.848          & {\textbf{0.690}} \\

			+LDS+ConR                                    & 7.16                                                 & 6.61                                                     & 7.97                                                 & 9.62           & 7.43           & 6.84           & 12.38          & 21.98          & 0.927           & 0.901          & 1.021          & 1.087            \\

			\textbf{+LDS+ACCon }                         & 6.89                                                 & \textbf{6.38}                                            & 7.55                                                 & 9.89           & \textbf{7.34}  & \textbf{6.76 } & 11.98          & 22.13          & \textbf{0.806}  & \textbf{0.821} & 0.742          & 0.732            \\

			\textbf{+LDS+ACCon+ConR}                     & \textbf{6.84}                                        & 6.41                                                     & \textbf{7.42}                                        & \textbf{9.26}  & \textbf{7.34}  & 6.79           & \textbf{11.78} & \textbf{21.91} & 0.820           & 0.846          & \textbf{0.715} & 0.692            \\

			\midrule
			Focal-R                                      & 7.64                                                 & 6.68                                                     & 9.22                                                 & 13.00          & 7.97           & 7.12           & 15.14          & 26.96          & 0.951           & 0.843          & 1.425          & 0.957            \\
			+LDS                                         & 7.56                                                 & 6.67                                                     & 8.82                                                 & 12.40          & 7.90           & 7.10           & 14.72          & 25.84          & 0.930           & 0.807          & 1.449          & 0.993            \\

			+LDS+NaïveSupCon                             & 7.67                                                 & 6.94                                                     & 8.87                                                 & 11.31          & 7.84           & 7.02           & 14.89          & 25.61          & 0.957           & 0.917          & 1.075          & 1.089            \\

			+LDS+AdaSupCon                               & 7.59                                                 & 6.82                                                     & 8.92                                                 & 11.18          & 7.80           & 7.06           & 14.02          & 25.12          & 1.056           & 0.932          & 1.477          & 1.358            \\

			+LDS+RankSIM                                 & 7.25                                                 & \textbf{6.40}                                            & 8.71                                                 & 11.24          & 7.71           & 6.99           & 13.65          & 25.97          & 0.887           & 0.889          & 0.918          & 0.745            \\

			+LDS+ConR                                    & 7.23                                                 & 6.63                                                     & 8.30                                                 & 11.89          & 7.85           & 7.01           & 14.31          & 25.23          & 0.944           & 0.912          & 1.102          & 0.998            \\

			\textbf{+LDS+ACCon}                          & 6.90                                                 & 6.43                                                     & 7.48                                                 & 9.74           & \textbf{7.38}  & \textbf{6.68}  & 13.12          & 24.43          & 0.801           & 0.822          & \textbf{0.717} & 0.702            \\

			\textbf{+LDS+ACCon+ConR}                     & \textbf{6.83}                                        & 6.44                                                     & \textbf{7.36}                                        & \textbf{9.02}  & 7.40           & 6.72           & \textbf{13.02} & \textbf{24.19} & \textbf{0.799}  & \textbf{0.819} & 0.722          & \textbf{0.681 }  \\

			\midrule   
			Ours vs. Vanilla                             & \textcolor{teal}{$\uparrow$12.4\%}                   & \textcolor{teal}{$\uparrow$5.44\%}
			                                             & \textcolor{teal}{$\uparrow$22.9\%}                   & \textcolor{teal}{$\uparrow$34.0\%}

			                                             & \textcolor{teal}{$\uparrow$10.8\%}                   & \textcolor{teal}{$\uparrow$8.58\%}
			                                             & \textcolor{teal}{$\uparrow$23.8\%}
			                                             & \textcolor{teal}{$\uparrow$17.8\%}

			                                             & \textcolor{teal}{$\uparrow$18.0\%}
			                                             & \textcolor{teal}{$\uparrow$4.58\%}
			                                             & \textcolor{teal}{$\uparrow$53.0\%}
			                                             & \textcolor{teal}{$\uparrow$30.8\%}                                                                                                                                                                                                                                                                                                 \\
			Ours vs. SOTA
			                                             & \textcolor{teal}{$\uparrow$2.58\%}
			                                             & \textcolor{teal}{$\uparrow$1.88\%}
			                                             & \textcolor{teal}{$\uparrow$6.60\%}
			                                             & \textcolor{teal}{$\uparrow$6.24\%}

			                                             & \textcolor{teal}{$\uparrow$1.91\%}
			                                             & \textcolor{teal}{$\uparrow$2.07\%}
			                                             & \textcolor{teal}{$\uparrow$3.92\%}
			                                             & \textcolor{teal}{$\uparrow$1.55\%}

			                                             & \textcolor{teal}{$\uparrow$8.48\%}
			                                             & \textcolor{teal}{$\uparrow$0.85\%}
			                                             & \textcolor{teal}{$\uparrow$6.78\%}
			                                             & \textcolor{teal}{$\uparrow$3.40\%}                                                                                                                                                                                                                                                                                                 \\
			\bottomrule   
		\end{tabular}}
    	\caption{Performance comparison with various imbalanced regression methods on the AgeDB-DIR, IMDB-WIKI-DIR, and STS-B-DIR.
		The Vanilla, LDS, SqrtINV, Focal-R, RankSim and ConR models are sourced from \cite{gong2022ranksim,yang2021delving,keramati2023conr}.
		The optimal outcomes for each method are highlighted in \textbf{bold font}.
	}
	\label{tab22}
	\vspace{-0.2em}
\end{table*}

\subsection{Performance on Limited Training Data}

Although massive datasets have been instrumental in advancing contemporary deep learning, the labeling of extensive datasets for training purposes is often impractical due to associated costs and time constraints, particularly in scientific AI applications.
Consequently, there is a growing demand for models that maintain robustness despite training data scarcity.
To address this, we evaluated our method's performance under limited training data conditions, thereby assessing its data efficiency.
As shown in Fig. \ref{data_efficency}, our method demonstrates enhanced robustness to limited datasets compared to vanilla and RankSim methods.

\subsection{Ablation Studies}

We conducted a series of ablation studies to analyze the components of our deep regression framework.
Initially, we compared the efficacy of using features extracted directly from $f_{\theta}\left(\cdot\right)$ versus features obtained after applying a projection layer for downstream tasks.
Although it is common practice to utilize features prior to the projection layer (denoted as \emph{Before Proj.}) for downstream tasks, our results, as presented in Appendix Table 1, demonstrate that leveraging features post-projection layer leads to superior performance.
Subsequently, we assessed the model's performance after removing the projection layer (denoted as \emph{w/o Proj.}).
The results prove the necessity of including the projection layer in the architecture.
Furthermore, we implemented a two-stage training scheme, a strategy commonly employed in self-supervised contrastive learning.
However, this two-stage scheme presented challenges in model convergence and yielded inferior results compared to training with a multi-task loss function.
Further analysis of feature representation, detailed in Appendix C, indicates that our framework's design is advantageous for enhancing feature representation.

We conducted additional ablation studies to assess the impact of hyperparameters, including the weight coefficient $\gamma$ and the representation dimension $z$.
Our method demonstrated robustness to variations in these parameters.
Comprehensive experimental results are presented in Appendix C.

\subsubsection{Qualitative Visualization of Feature Space:}

We employed t-SNE to visualize the extracted feature representations of our method and the compared baselines on the AgeDB-Natural test dataset, as illustrated in Fig. \ref{fig4} (a).
Notably, the representations extracted by our method exhibit continuity, compactness, and relative symmetry in low-dimensional space, corresponding well to the label space.
Specifically, RankSIM, BMSE, and our method demonstrate the capability to obtain continuous and compact features in low-dimensional space, consistent with their design to preserve label relationships.

\begin{figure}[t]
	\centering
	\includegraphics[width=8.3cm]{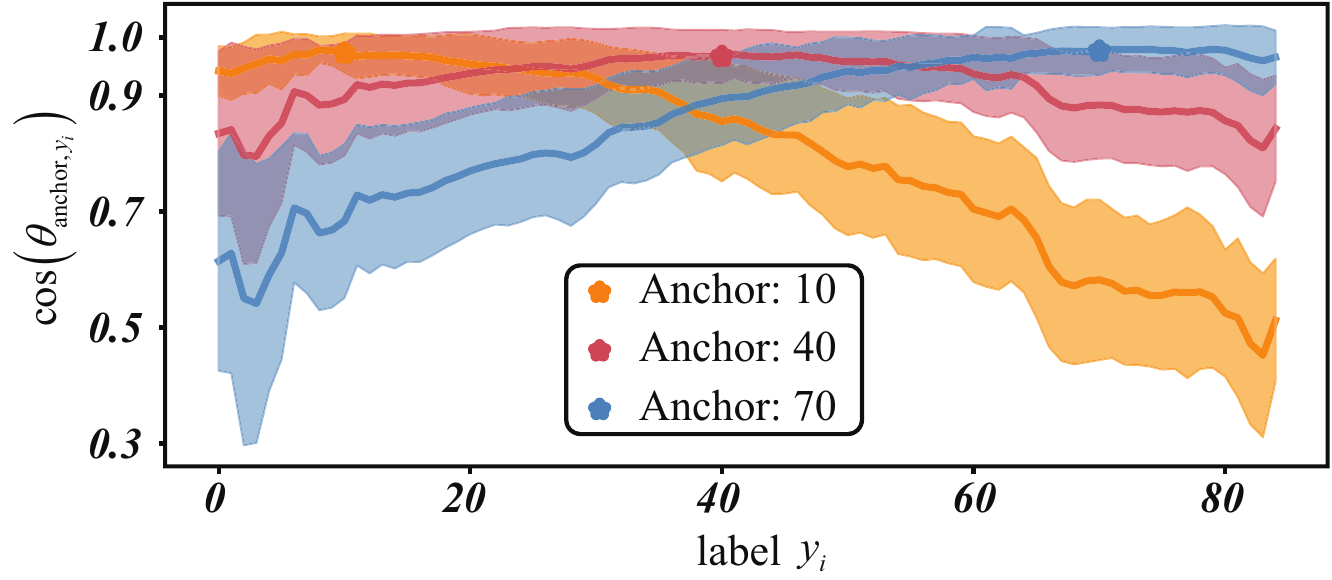}
	
	\caption{The cosine similarity changes as the label distance varies between the anchor and contrast samples in the AgeDB-Natural test dataset.
		The stars represent the anchors, while the plot depicts the mean cosine similarity between each anchor and all its contrastive samples.
		The shaded region indicates the standard deviation of these cosine similarities.}
	\label{plot_distance}
	\vspace{-0.4cm}
\end{figure}

\subsubsection{Correlation Between Feature Representation Similarity and Label Distance}

To further investigate the learned representations, we present a joint distribution analysis of $\cos\left(\theta_{i,j}\right)$ and the normalized label distance $|y_{i}-y_{j}|/100$ on the AgeDB-Natural test dataset, as shown in Fig. \ref{fig4} (b).
The extracted representations by our proposed method exhibit a negative correlation with the normalized label distance, achieving a Pearson correlation of -0.735.
This indicates that our method effectively captures feature representations with inherent label ordering information, aligning with our initial motivation.
The ideal cosine distance among feature representations should be distributed in the range $\left[-1, 1\right]$, whereas in our method, it mostly falls within $\left[0.3, 1\right]$.
We attribute this deviation to label imbalance, a known challenge in supervised contrastive learning for image recognition, as discussed in previous studies \cite{zhu2022balanced,li2022targeted}.
Fig. \ref{plot_distance} illustrates the changes in cosine similarity as the label distance varies between anchor and contrast samples on the AgeDB-Natural test dataset.
This analysis highlights the alignment between label-space neighbors and feature-space neighbors, supporting the rationale behind our proposed approach.
Notably, we observed increased standard variance as the distance grows, indicating challenges in performance improvement, particularly in the few-shot region.
In summary, this visualization provides evidence that our ACCon contributes to the model's ability to learn feature representations that preserve label-space neighbor information.

\section{Conclusion}

Although continual, long-tail, imbalanced, and even missing data in regression targets complicate the regression task, the additional information embedded within these targets offers valuable insights to enhance regression performance.
However, current methods fall short of capturing sufficient relationship information.
For instance, order-aware approaches overlook the distance between labels, while existing distance-aware methods struggle to ensure that representation similarities accurately reflect label distances.
To address this, we propose a hypothesis of a linear negative correlation between feature similarities and label distances, implementing this through an angle-compensated supervised contrastive regularizer.
Through extensive experiments on age estimation and semantic textual similarity tasks, we show that our method not only surpasses several state-of-the-art approaches but also excels in terms of data efficiency and performance on imbalanced datasets.
Furthermore, additional analysis and theoretical validation confirm that our method aligns well with its underlying motivation, supporting the reasonableness of our assumption and the effectiveness of our approach.
Our method offers promising potential for various applications in deep regression, contributing to the advancement of this field.

\section{Acknowledgments}

This study was supported by the Key Research and Development Program of Guangdong Province (grant No. 2021B0101400003) and the corresponding author is Jianzong Wang (jzwang@188.com).

\bibliography{refs}

\begin{thebibliography}{34}
\providecommand{\natexlab}[1]{#1}

\bibitem[{Belagiannis et~al.(2015)Belagiannis, Rupprecht, Carneiro, and Navab}]{belagiannis2015robust}
Belagiannis, V.; Rupprecht, C.; Carneiro, G.; and Navab, N. 2015.
\newblock Robust optimization for deep regression.
\newblock In \emph{Proceedings of the IEEE international conference on computer vision}, 2830--2838.

\bibitem[{Cer et~al.(2017)Cer, Diab, Agirre, Lopez-Gazpio, and Specia}]{cer2017semeval}
Cer, D.; Diab, M.; Agirre, E.; Lopez-Gazpio, I.; and Specia, L. 2017.
\newblock SemEval-2017 Task 1: Semantic Textual Similarity Multilingual and Crosslingual Focused Evaluation.
\newblock In \emph{Proceedings of the 11th International Workshop on Semantic Evaluation (SemEval-2017)}. Association for Computational Linguistics.

\bibitem[{Chandrasekaran and Mago(2021)}]{chandrasekaran2021evolution}
Chandrasekaran, D.; and Mago, V. 2021.
\newblock Evolution of semantic similarity—a survey.
\newblock \emph{ACM Computing Surveys (CSUR)}, 54(2): 1--37.

\bibitem[{Chen et~al.(2020)Chen, Kornblith, Norouzi, and Hinton}]{chen2020simple}
Chen, T.; Kornblith, S.; Norouzi, M.; and Hinton, G. 2020.
\newblock A simple framework for contrastive learning of visual representations.
\newblock In \emph{International conference on machine learning}, 1597--1607. PMLR.

\bibitem[{Chen, Ma, and Lin(2021)}]{chen2021time}
Chen, Z.; Ma, Q.; and Lin, Z. 2021.
\newblock Time-Aware Multi-Scale RNNs for Time Series Modeling.
\newblock In \emph{IJCAI}, 2285--2291.

\bibitem[{Dai et~al.(2021)Dai, Li, Chiu, Kuo, and Cheng}]{dai2021adaptive}
Dai, W.; Li, X.; Chiu, W. H.~K.; Kuo, M.~D.; and Cheng, K.-T. 2021.
\newblock Adaptive contrast for image regression in computer-aided disease assessment.
\newblock \emph{IEEE Transactions on Medical Imaging}, 41(5): 1255--1268.

\bibitem[{Ermolov et~al.(2021)Ermolov, Siarohin, Sangineto, and Sebe}]{ermolov2021whitening}
Ermolov, A.; Siarohin, A.; Sangineto, E.; and Sebe, N. 2021.
\newblock Whitening for self-supervised representation learning.
\newblock In \emph{International Conference on Machine Learning}, 3015--3024. PMLR.

\bibitem[{Gao et~al.(2017)Gao, Xing, Xie, Wu, and Geng}]{gao2017deep}
Gao, B.-B.; Xing, C.; Xie, C.-W.; Wu, J.; and Geng, X. 2017.
\newblock Deep label distribution learning with label ambiguity.
\newblock \emph{IEEE Transactions on Image Processing}, 26(6): 2825--2838.

\bibitem[{Gong, Mori, and Tung(2022)}]{gong2022ranksim}
Gong, Y.; Mori, G.; and Tung, F. 2022.
\newblock RankSim: Ranking Similarity Regularization for Deep Imbalanced Regression.
\newblock In \emph{International Conference on Machine Learning}, 7634--7649. PMLR.

\bibitem[{Graf et~al.(2021)Graf, Hofer, Niethammer, and Kwitt}]{graf2021dissecting}
Graf, F.; Hofer, C.; Niethammer, M.; and Kwitt, R. 2021.
\newblock Dissecting supervised contrastive learning.
\newblock In \emph{International Conference on Machine Learning}, 3821--3830. PMLR.

\bibitem[{Huber(1992)}]{huber1992robust}
Huber, P.~J. 1992.
\newblock Robust estimation of a location parameter.
\newblock In \emph{Breakthroughs in statistics: Methodology and distribution}, 492--518. Springer.

\bibitem[{Jiang et~al.(2020)Jiang, Huang, Geng, and Deng}]{jiang2020multi}
Jiang, W.; Huang, K.; Geng, J.; and Deng, X. 2020.
\newblock Multi-scale metric learning for few-shot learning.
\newblock \emph{IEEE Transactions on Circuits and Systems for Video Technology}, 31(3): 1091--1102.

\bibitem[{Keramati, Meng, and Evans(2024)}]{keramati2023conr}
Keramati, M.; Meng, L.; and Evans, R.~D. 2024.
\newblock Conr: Contrastive regularizer for deep imbalanced regression.

\bibitem[{Khosla et~al.(2020)Khosla, Teterwak, Wang, Sarna, Tian, Isola, Maschinot, Liu, and Krishnan}]{khosla2020supervised}
Khosla, P.; Teterwak, P.; Wang, C.; Sarna, A.; Tian, Y.; Isola, P.; Maschinot, A.; Liu, C.; and Krishnan, D. 2020.
\newblock Supervised contrastive learning.
\newblock \emph{Advances in neural information processing systems}, 33: 18661--18673.

\bibitem[{Lathuili{\`e}re et~al.(2019)Lathuili{\`e}re, Mesejo, Alameda-Pineda, and Horaud}]{lathuiliere2019comprehensive}
Lathuili{\`e}re, S.; Mesejo, P.; Alameda-Pineda, X.; and Horaud, R. 2019.
\newblock A comprehensive analysis of deep regression.
\newblock \emph{IEEE transactions on pattern analysis and machine intelligence}, 42(9): 2065--2081.

\bibitem[{Lee et~al.(2021)Lee, Lee, Kim, Yi, and Kim}]{lee2021patch}
Lee, S.; Lee, J.; Kim, B.; Yi, E.; and Kim, J. 2021.
\newblock Patch-wise attention network for monocular depth estimation.
\newblock In \emph{Proceedings of the AAAI Conference on Artificial Intelligence}, volume~35, 1873--1881.

\bibitem[{Li et~al.(2022)Li, Cao, Yuan, Fan, Yang, Feris, Indyk, and Katabi}]{li2022targeted}
Li, T.; Cao, P.; Yuan, Y.; Fan, L.; Yang, Y.; Feris, R.~S.; Indyk, P.; and Katabi, D. 2022.
\newblock Targeted supervised contrastive learning for long-tailed recognition.
\newblock In \emph{Proceedings of the IEEE/CVF Conference on Computer Vision and Pattern Recognition}, 6918--6928.

\bibitem[{Li et~al.(2021)Li, Huang, Lu, Feng, and Zhou}]{li2021learning}
Li, W.; Huang, X.; Lu, J.; Feng, J.; and Zhou, J. 2021.
\newblock Learning probabilistic ordinal embeddings for uncertainty-aware regression.
\newblock In \emph{Proceedings of the IEEE/CVF conference on computer vision and pattern recognition}, 13896--13905.

\bibitem[{Liu et~al.(2023)Liu, Zhu, Shen, Liu, Razavian, Geras, and Fernandez-Granda}]{liu2023multiple}
Liu, K.; Zhu, W.; Shen, Y.; Liu, S.; Razavian, N.; Geras, K.~J.; and Fernandez-Granda, C. 2023.
\newblock Multiple instance learning via iterative self-paced supervised contrastive learning.
\newblock In \emph{Proceedings of the IEEE/CVF Conference on Computer Vision and Pattern Recognition}, 3355--3365.

\bibitem[{Moschoglou et~al.(2017)Moschoglou, Papaioannou, Sagonas, Deng, Kotsia, and Zafeiriou}]{moschoglou2017agedb}
Moschoglou, S.; Papaioannou, A.; Sagonas, C.; Deng, J.; Kotsia, I.; and Zafeiriou, S. 2017.
\newblock Agedb: the first manually collected, in-the-wild age database.
\newblock In \emph{proceedings of the IEEE conference on computer vision and pattern recognition workshops}, 51--59.

\bibitem[{Pan et~al.(2018)Pan, Han, Shan, and Chen}]{pan2018mean}
Pan, H.; Han, H.; Shan, S.; and Chen, X. 2018.
\newblock Mean-variance loss for deep age estimation from a face.
\newblock In \emph{Proceedings of the IEEE conference on computer vision and pattern recognition}, 5285--5294.

\bibitem[{Ren et~al.(2022)Ren, Zhang, Yu, and Liu}]{ren2022balanced}
Ren, J.; Zhang, M.; Yu, C.; and Liu, Z. 2022.
\newblock Balanced mse for imbalanced visual regression.
\newblock In \emph{Proceedings of the IEEE/CVF Conference on Computer Vision and Pattern Recognition}, 7926--7935.

\bibitem[{Rogez, Weinzaepfel, and Schmid(2017)}]{rogez2017lcr}
Rogez, G.; Weinzaepfel, P.; and Schmid, C. 2017.
\newblock Lcr-net: Localization-classification-regression for human pose.
\newblock In \emph{Proceedings of the IEEE Conference on Computer Vision and Pattern Recognition}, 3433--3441.

\bibitem[{Rothe, Timofte, and Van~Gool(2018)}]{rothe2018deep}
Rothe, R.; Timofte, R.; and Van~Gool, L. 2018.
\newblock Deep expectation of real and apparent age from a single image without facial landmarks.
\newblock \emph{International Journal of Computer Vision}, 126(2-4): 144--157.

\bibitem[{Seifi et~al.(2024)Seifi, Reino, Chumerin, and Aljundi}]{seifi2024ood}
Seifi, S.; Reino, D.~O.; Chumerin, N.; and Aljundi, R. 2024.
\newblock OOD Aware Supervised Contrastive Learning.
\newblock In \emph{Proceedings of the IEEE/CVF Winter Conference on Applications of Computer Vision}, 1956--1966.

\bibitem[{Shi, Cao, and Raschka(2023)}]{shi2023deep}
Shi, X.; Cao, W.; and Raschka, S. 2023.
\newblock Deep neural networks for rank-consistent ordinal regression based on conditional probabilities.
\newblock \emph{Pattern Analysis and Applications}, 26(3): 941--955.

\bibitem[{Sun et~al.(2023)Sun, Shi, Gao, Ren, de~Rijke, and Ren}]{sun2023contrastive}
Sun, W.; Shi, Z.; Gao, S.; Ren, P.; de~Rijke, M.; and Ren, Z. 2023.
\newblock Contrastive learning reduces hallucination in conversations.
\newblock In \emph{Proceedings of the AAAI Conference on Artificial Intelligence}, volume~37, 13618--13626.

\bibitem[{Tabelini et~al.(2021)Tabelini, Berriel, Paixao, Badue, De~Souza, and Oliveira-Santos}]{tabelini2021polylanenet}
Tabelini, L.; Berriel, R.; Paixao, T.~M.; Badue, C.; De~Souza, A.~F.; and Oliveira-Santos, T. 2021.
\newblock Polylanenet: Lane estimation via deep polynomial regression.
\newblock In \emph{2020 25th International Conference on Pattern Recognition (ICPR)}, 6150--6156. IEEE.

\bibitem[{Wang et~al.(2018)Wang, Singh, Michael, Hill, Levy, and Bowman}]{wang2018glue}
Wang, A.; Singh, A.; Michael, J.; Hill, F.; Levy, O.; and Bowman, S.~R. 2018.
\newblock GLUE: A Multi-Task Benchmark and Analysis Platform for Natural Language Understanding.
\newblock \emph{EMNLP 2018}, 353.

\bibitem[{Wang, Sanchez, and Li(2022)}]{wang2022improving}
Wang, H.; Sanchez, V.; and Li, C.-T. 2022.
\newblock Improving face-based age estimation with attention-based dynamic patch fusion.
\newblock \emph{IEEE Transactions on Image Processing}, 31: 1084--1096.

\bibitem[{Yang et~al.(2021)Yang, Zha, Chen, Wang, and Katabi}]{yang2021delving}
Yang, Y.; Zha, K.; Chen, Y.; Wang, H.; and Katabi, D. 2021.
\newblock Delving into deep imbalanced regression.
\newblock In \emph{International Conference on Machine Learning}, 11842--11851. PMLR.

\bibitem[{Zha et~al.(2024)Zha, Cao, Son, Yang, and Katabi}]{zha2024rank}
Zha, K.; Cao, P.; Son, J.; Yang, Y.; and Katabi, D. 2024.
\newblock Rank-N-Contrast: Learning Continuous Representations for Regression.
\newblock \emph{Advances in Neural Information Processing Systems}, 36.

\bibitem[{Zhang et~al.(2022)Zhang, Yang, Mi, Zheng, and Yao}]{zhang2022improving}
Zhang, S.; Yang, L.; Mi, M.~B.; Zheng, X.; and Yao, A. 2022.
\newblock Improving Deep Regression with Ordinal Entropy.
\newblock In \emph{The Eleventh International Conference on Learning Representations}.

\bibitem[{Zhu et~al.(2022)Zhu, Wang, Chen, Chen, and Jiang}]{zhu2022balanced}
Zhu, J.; Wang, Z.; Chen, J.; Chen, Y.-P.~P.; and Jiang, Y.-G. 2022.
\newblock Balanced contrastive learning for long-tailed visual recognition.
\newblock In \emph{Proceedings of the IEEE/CVF Conference on Computer Vision and Pattern Recognition}, 6908--6917.

\end{thebibliography}
\appendix


\end{document}


\maketitle

\section{Appendix A. Proof}
\subsection{A1.The Derivation of $\cos\left(\tilde{\theta}_{i,m}\right)$}
\label{supply:a1}
Given the assumption of ACCon, we assume that the cosine similarity between anchor and negative pairs as follows:
\begin{equation}
    \hat{\theta} = \frac{y_{\text{neg}} - y_{\text{anc}}}{\max\left(\mathcal{Y}\right)-\min\left(\mathcal{Y}\right)} \pi,
\end{equation}

In our approach, we modify the standard supervised contrastive loss within a mini-batch.
Typically, this loss constrains the representations of anchors and negatives to be as far apart as possible, which is equivalent to setting the included angle $\tilde{\theta}_{i,m}$ to $\pi$.
Based on this principle, we construct the following equation:

\begin{equation}
    \cos\left(\tilde{\theta}_{i,m}\right) = \cos\left(\hat{\theta}_{i,m} + \pi - \frac{y_{m} - y_{i}}{\max\left(\mathcal{Y}\right)-\min\left(\mathcal{Y}\right)}\pi\right).
\end{equation}
where $\hat{\theta}$ is the ideal cosine similarity.
Besides, we define the compensated angle $\varphi$, which is formulated as:
\begin{equation}
    \varphi = \pi\left(1 -\frac{y_{m} - y_{i}}{\max\left(\mathcal{Y}\right)-\min\left(\mathcal{Y}\right)}\right).
\end{equation}
Then:
\begin{equation}
    \begin{aligned}
        \cos\left(\tilde{\theta}_{i,m}\right) & = \cos\left(\hat{\theta}_{i,m} + \varphi_{i,m}\right)                                                                                      \\
                                              & = \cos\left(\hat{\theta}_{i,m}\right)\cos\left(\varphi_{i,m}\right) - \sin\left(\hat{\theta}_{i,m}\right)\sin(\varphi)                     \\
                                              & = \cos\left(\hat{\theta}_{i,m}\right)\cos\left(\varphi\right) \pm \sin\left(\varphi\right)\sqrt{1-\cos\left(\hat{\theta}_{i,m}\right)^{2}}
    \end{aligned}
\end{equation}
Because we have supposed the optimized $\tilde{\theta}_{i,m}=\hat{\theta}_{i,m} + \varphi=\pi$, the above formulation could be approximated to:
\begin{equation}
    \begin{aligned}
        \cos\left(\tilde{\theta}_{i,m}\right) = \cos\left(\hat{\theta}_{i,m}\right)\cos\left(\varphi\right) - \left|\sin\left(\varphi\right)\right|\sqrt{1-\cos\left(\hat{\theta}_{i,m}\right)^{2}}
    \end{aligned}
\end{equation}
In the deep regression framework, we have:
\begin{equation}
    \cos\left(\hat{\theta}_{i,m}\right) = z_{i}z_{m}^{T}
\end{equation}
where $ z_{i}, z_{m}$ are L2-normalized feature representations. Then, we have:
\begin{equation}
    \cos(\tilde{\theta}_{i,m}) = z_{i}z_{m}^{T}cos(\varphi) - |sin(\varphi)|\sqrt{1-(z_{i}z_{m}^{T})^{2}}
\end{equation}
The term $1-\left(z_{i}z_{m}^{T}\right)^{2}$ may approach 0 during the training process.
Therefore, we add a samll smoothing term $\epsilon$.
Then, the final $\cos\left(\tilde{\theta}_{i,m}\right)$ becomes:
\begin{equation}
    \cos\left(\tilde{\theta}_{i,m}\right) = z_{i}z_{m}^{T}\cos\left(\varphi\right) - |\sin\left(\varphi\right)|\sqrt{1-\left(z_{i}z_{m}^{T}\right)^{2}+\epsilon}
\end{equation}

\subsection{A2. Proof of Theorem 1}
\label{supply:a2}

Next, we will give a prove of the lower bound $L^{*}$ of $\mathcal{L}^{\mathrm{acCON}}$:
\begin{equation*}
    L^{*} =  \frac{1}{4N^{2}} \sum_{i=0}^{2N} \sum\limits_{m \in \mathcal{N}\left(i\right)} \cos\left(\tilde{\theta}_{i,m}\right)/\tau - \frac{\mathrm{log} N / \tau}{2N}.
\end{equation*}

For each anchor $x_{i}$, we have:
\begin{equation}
    \scriptsize
    \begin{aligned}
        \mathcal{L}_{i}^{\mathrm{ac}}
         & =-\frac{1}{|\mathcal{P}(i)|}\sum\limits_{p\in \mathcal{P}(i)}{\log }\frac{\exp \left( {{z}_{i}}z_{p}^{T}/\tau  \right)}{\sum\limits_{k\in \mathcal{P}(i)}{\exp }\left( {{z}_{i}}z_{k}^{T}/\tau  \right)+\sum\limits_{m\in \mathcal{N}(i)}{\exp }\left( \cos \left( {{{\tilde{\theta }}}_{i,m}} \right)/\tau  \right)}                                                                                                                                                                           \\
         & =\frac{1}{|\mathcal{P}\left( i \right)|}\sum\limits_{p\in \mathcal{P}\left( i \right)}{\log }\left[ \frac{\sum\limits_{k\in \mathcal{P}\left( i \right)}{\exp }\left( {{z}_{i}}z_{k}^{T}/\tau  \right)}{\exp \left( {{z}_{i}}z_{p}^{T}/\tau  \right)}+\frac{\sum\limits_{m\in \mathcal{N}\left( i \right)}{\exp }\left( \cos \left( {{\theta }_{i,m}} \right)/\tau  \right)}{\exp \left( {{z}_{i}}z_{p}^{T}/\tau  \right)} \right]                                                              \\
         & \ge \frac{1}{|\mathcal{P}\left( i \right)|}\sum\limits_{p\in \mathcal{P}\left( i \right)}{\log }\left[ \frac{\sum\limits_{k\in \mathcal{P}\left( i \right)}{\exp }\left( {{z}_{i}}z_{k}^{T}/\tau  \right)}{\sum\limits_{k\in \mathcal{P}\left( i \right)}{\exp }\left( {{z}_{i}}z_{k}^{T}/\tau  \right)}+\frac{\sum\limits_{m\in \mathcal{N}\left( i \right)}{\exp \left( \cos \left( {{{\tilde{\theta }}}_{i,m}} \right)/\tau  \right)}}{\exp \left( {{z}_{i}}z_{p}^{T}/\tau  \right)} \right] \\
         & =\frac{1}{|\mathcal{P}\left( i \right)|}\sum\limits_{p\in \mathcal{P}\left( i \right)}{\log }\left[ 1+\frac{\sum\limits_{m\in \mathcal{N}\left( i \right)}{\exp }\left( \cos \left( {{{\tilde{\theta }}}_{i,m}} \right)/\tau  \right)}{\exp \left( {{z}_{i}}z_{p}^{T}/\tau  \right)} \right]                                                                                                                                                                                                    \\
         & \ge \frac{1}{|\mathcal{P}\left( i \right)|}\sum\limits_{p\in \mathcal{P}\left( i \right)}{\log }\left[ \frac{\sum\limits_{m\in \mathcal{N}\left( i \right)}{\exp }\left( \cos \left( {{{\tilde{\theta }}}_{i,m}} \right)/\tau  \right)}{\exp \left( {{z}_{i}}z_{p}^{T}/\tau  \right)} \right]                                                                                                                                                                                                   \\
         & \ge \frac{1}{|\mathcal{P}\left( i \right)|}\left[ \log \sum\limits_{m\in \mathcal{N}\left( i \right)}{\exp }\left( \cos \left( {{{\tilde{\theta }}}_{i,m}} \right)/\tau  \right)-\log \sum\limits_{p\in \mathcal{P}\left( i \right)}{\exp }\left( {{z}_{i}}z_{p}^{T}/\tau  \right) \right]
    \end{aligned}
\end{equation}

\onecolumn

Therefore, we have:
\begin{equation}
    \begin{aligned}
        {\mathcal{L}}_{\text{ACCon}}
         & =\frac{1}{2N}\sum\limits_{i=0}^{2N}{\mathcal{L}_{i}^{\mathrm{ac}}}                                                                                                                                                                                                                                                                                 \\
         & \ge \frac{1}{2N}\sum\limits_{i=0}^{2N}{\frac{1}{|\mathcal{P}\left( i \right)|}}\left[ \text{log}\sum\limits_{m\in \mathcal{N}\left( i \right)}{\text{exp}}\left( \cos \left( {{{\tilde{\theta }}}_{i,m}} \right)/\tau  \right)-\text{log}\sum\limits_{p\in \mathcal{P}\left( i \right)}{\text{exp}}\left( {{z}_{i}}z_{p}^{T}/\tau  \right) \right] \\
         & =\frac{1}{2N}\sum\limits_{i=0}^{2N}{\frac{1}{2N}}\left[ \text{log}\sum\limits_{m\in \mathcal{N}\left( i \right)}{\text{exp}}\left( \cos \left( {{{\tilde{\theta }}}_{i,m}} \right)/\tau  \right)-\text{log}\sum\limits_{p\in \mathcal{P}\left( i \right)}{\text{exp}}\left( {{z}_{i}}z_{p}^{T}/\tau  \right) \right]                               \\
         & =\frac{1}{4{{N}^{2}}}\sum\limits_{i=0}^{2N}{\left[ \text{log}\sum\limits_{m\in \mathcal{N}\left( i \right)}{\text{exp}}\left( \cos \left( {{{\tilde{\theta }}}_{i,m}} \right)/\tau  \right)-\text{log}\sum\limits_{p\in \mathcal{P}\left( i \right)}{\text{exp}}\left( {{z}_{i}}z_{p}^{T}/\tau  \right) \right]}                                   \\
         & \ge \frac{1}{4{{N}^{2}}}\left[ \sum\limits_{i=0}^{2N}{\text{log}}\sum\limits_{m\in \mathcal{N}\left( i \right)}{\text{exp}}\left( \cos \left( {{{\tilde{\theta }}}_{i,m}} \right)/\tau  \right)-\sum\limits_{i=0}^{2N}{\text{log}}\sum\limits_{p\in \mathcal{P}\left( i \right)}{\text{exp}}\left( {{z}_{i}}z_{p}^{T}/\tau  \right) \right].
    \end{aligned}
\end{equation}

Considering the $\mathrm{exp}(z_{i}z_{p}^{T} / \tau) \leq \frac{1}{\tau}$, we have:
\begin{equation}
    \begin{aligned}
        \mathcal{L}_{\text{ACCon}} & \geq \frac{1}{4N^{2}} \left[ \sum_{i=0}^{2N}  \mathrm{log}\sum\limits_{m \in \mathcal{N}\left(i\right)} \mathrm{exp}\left(\cos\left(\tilde{\theta}_{i,m}\right)/\tau\right) -
        \sum_{i=0}^{2N} \mathrm{log} \sum_{p \in \mathcal{P}\left(i\right)} \left(1 / \tau\right) \right]                                                                                                          \\
                                   & = \frac{1}{4N^{2}} \left[ \sum_{i=0}^{2N}  \mathrm{log}\sum\limits_{m \in \mathcal{N}\left(i\right)} \mathrm{exp}\left(\cos\left(\tilde{\theta}_{i,m}\right)/\tau\right) -
        2N \mathrm{log} \left(|\mathcal{P}\left(i\right) | / \tau\right) \right]                                                                                                                                   \\
                                   & \geq \frac{1}{4N^{2}} \left[ \sum_{i=0}^{2N} \sum\limits_{m \in \mathcal{N}\left(i\right)} \cos\left(\tilde{\theta}_{i,m}\right)/\tau -
        2N \mathrm{log} \left(|\mathcal{P}\left(i\right) | / \tau\right) \right]                                                                                                                                   \\
                                   & \geq \frac{1}{4N^{2}} \left[ \sum_{i=0}^{2N} \sum\limits_{m \in \mathcal{N}\left(i\right)} \cos\left(\tilde{\theta}_{i,m}\right)/\tau -
            2N \mathrm{log} \left(2N / \tau\right) \right].
    \end{aligned}
\end{equation}

Then, we have:
\begin{equation}
    \begin{aligned}
        \mathcal{L}_{\text{ACCon}}  \geq \frac{1}{4N^{2}} \sum_{i=0}^{2N} \sum\limits_{m \in \mathcal{N}\left(i\right)} \cos\left(\tilde{\theta}_{i,m}\right)/\tau - \frac{\mathrm{log} N / \tau}{2N}
    \end{aligned}
\end{equation}


For $\mathcal{L}^{*} = \frac{1}{4N^{2}} \sum_{i=0}^{2N} \sum\limits_{m \in \mathcal{N}\left(i\right)} \cos\left(\tilde{\theta}_{i,m}\right)/\tau - \frac{\mathrm{log N / \tau}}{2N}$, we have:
\begin{equation}
    \begin{aligned}
        \arg \min_{\hat{\theta}_{i,m}} \left(\mathcal{L}^{*}\right) & = \arg \min _{\hat{\theta}_{i,m}} \frac{1}{4N^{2}} \sum_{i=0}^{2N} \sum\limits_{m \in \mathcal{N}\left(i\right)} \cos\left(\tilde{\theta}_{i,m}\right)/\tau                  \\
                                                                    & = \arg \min _{\hat{\theta}_{i,m}} \frac{1}{4N^{2}} \sum_{i=0}^{2N} \sum\limits_{m \in \mathcal{N}\left(i\right)}  \cos\left(\hat{\theta}_{i,m} + \varphi_{i,m}\right)/\tau .
    \end{aligned}
\end{equation}

Optimization of the above formulas leads to $\hat{\theta}_{i,m} + \varphi_{i,m} = \pi$ for each pair, which is equal to constraint the $\hat{\theta}_{i,m} = \frac{y_{m} - y_{i}}{\max(\mathcal{Y})-\min(\mathcal{Y})} \pi $.

\twocolumn

\section{Appendix B. Experiments Details}
\label{Experimentsdetails}

\subsection{B1. Label Distribution of Different Datasets}

Six datasets were used for our experiments.
For AgbDB-DIR, STS-B-DIR, and IMDB-WIKI-DIR, we directly follow the deep imbalanced regression benchmark proposed by \cite{yang2021delving}.
The AgbDB-Natural, STS-B-Natural, and IMDB-WIKI-Natural datasets were created by random sampling from their respective DIR datasets, maintaining consistent data splitting proportions with the DIR benchmark.

\begin{figure*}[t]
    \centering
    \includegraphics[width=16.5cm]{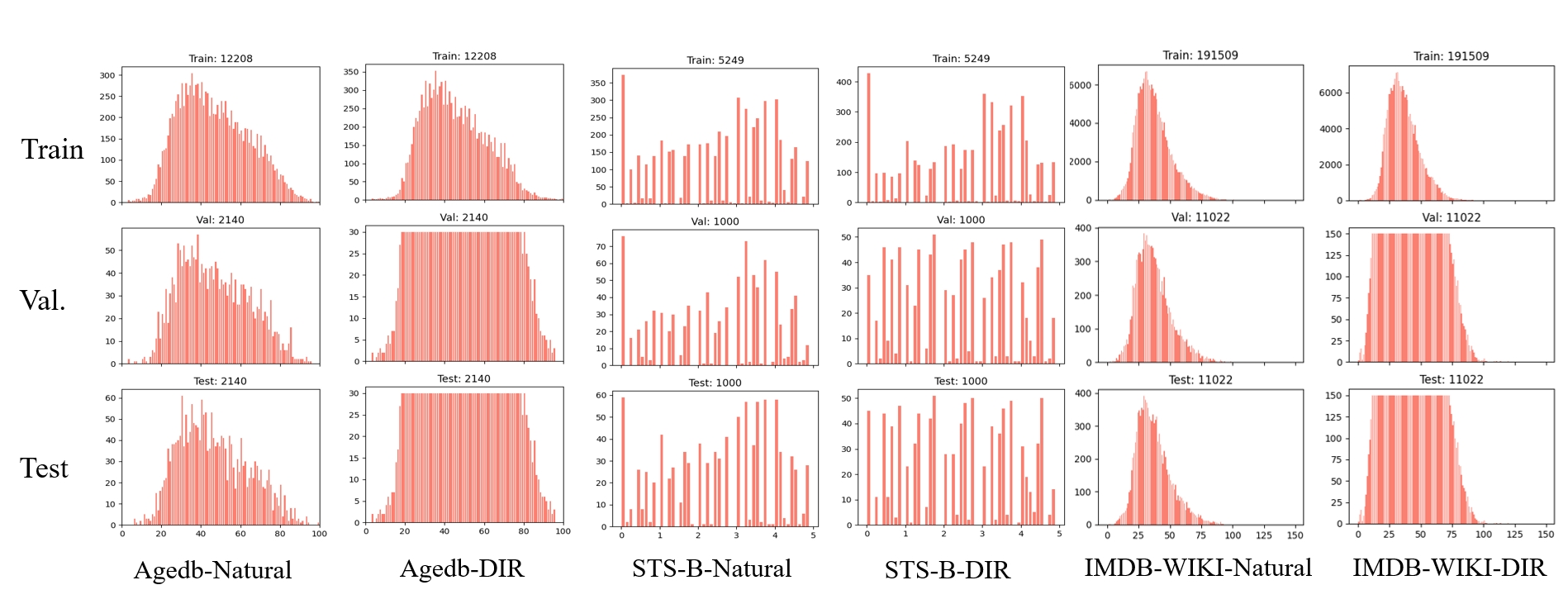}
    \caption{Label distributions for the training, testing, and validation datasets in AgbDB-Natural (DIR), STS-B-Natural (DIR), and IMDB-WIKI-Natural (DIR). In this context, \emph{Natural} signifies the random sampling of validation and test datasets, while \emph{DIR} indicates an imbalanced training dataset with a balanced validation/test dataset.}
    \label{label_distribution}
\end{figure*}

\begin{figure*}[h]
    \centering
    \begin{minipage}[t]{0.49\textwidth}
        \centering
        \centerline{\includegraphics[width=8.5cm]{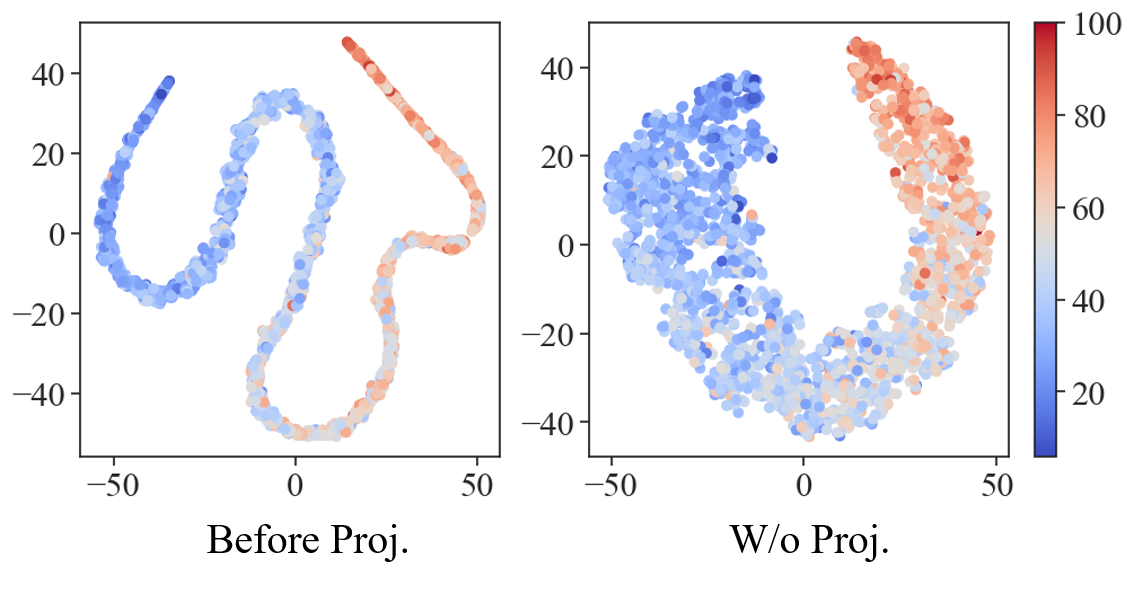}}
        \centerline{(a) t-SNE of feature representations} \medskip
    \end{minipage}
    \begin{minipage}[t]{0.49\textwidth}
        \centering
        \centerline{\includegraphics[width=8.5cm]{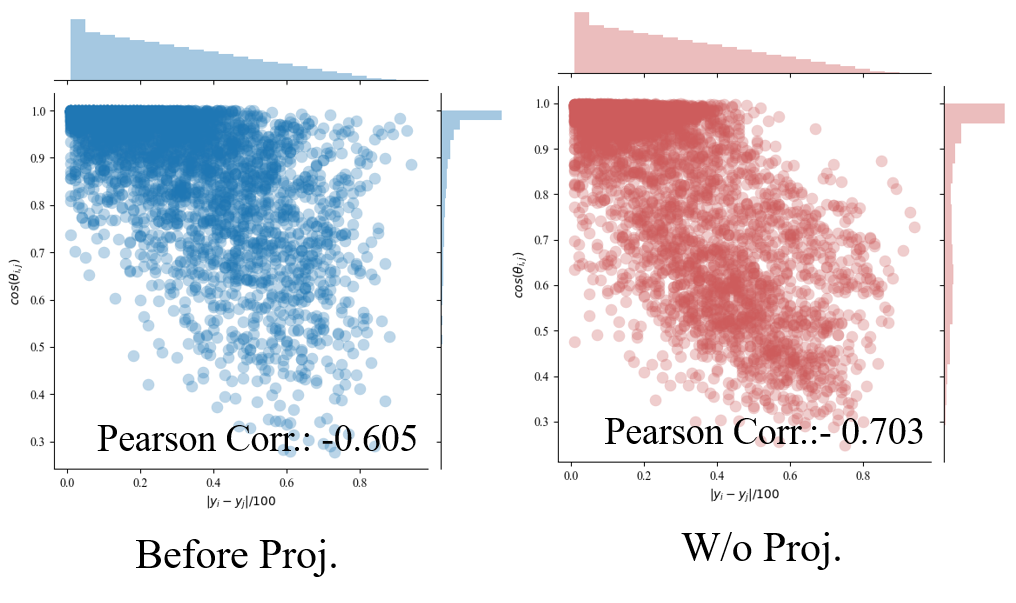}}
        \centerline{(b) Joint distribution}
    \end{minipage}
    \caption{The visualization and quantitation analysis of feature representations. The subfigure (a) is the t-SNE visualization of feature space on the AgeDB-natural test dataset. The subfigure (b) depicts the joint distribution of $\cos\left(\theta_{i,j}\right)$ and the labels distance $|y_{i}-y_{j}|/100$ on the AgeDB-natural test dataset using kernel density estimation.}
    \label{figS1}
\end{figure*}

\subsection{B2. AgeDB-Natural/DIR Training Protocol}

We trained our proposed model on AgeDB-Natural for 90 epochs using the Adam optimizer with a batch size of 64 on a single Tesla P100 GPU.
The initial learning rate was 0.00025, decaying by 0.0001 after the 60th epoch.
We set the temperature parameter $\tau$ to 0.05, weight coefficient $\gamma$ to 1.0, and the smoothing term to 1e-6.
Data preprocessing included random cropping, resizing to $224 \times 224$, and random horizontal flipping.
The AgeDB-DIR protocol was similar, with the key difference being a projection dimension of 128 for $g_{\phi}\left(\cdot\right)$, compared to 512 in AgeDB-Natural.

\subsection{B3. IMDB-WIKI-Natural/DIR Training Protocol}
The IMDB-WIKI-Natural training protocol is almost the same as AgeDB, including hyperparameters and preprocessing steps.
The IMDB-WIKI-DIR's projection dimension of $g_{\phi}(\cdot)$ is 128, and IMDB-WIKI-Natural is 512.

\subsection{B4. STS-B-Natural/DIR Training Protocol}

Following \cite{yang2021delving}, we trained on STS-B-Natural using the Adam optimizer with a learning rate of 1e-4 and a batch size of 128.
Validation occurred every 400 iterations, with training termination if validation error did not decrease after 30 consecutive checks (maximum 100 checks allowed).
For CCon hyperparameters on STS-B-Natural, we set the temperature parameter $\tau$ to 0.05, the weight coefficient $\gamma$ to 10, projection dimension to 2000, and the smoothing term to 1e-6.
STS-B-DIR training was similar, except for setting $\gamma$ adjusted to 0.1.

\begin{table*}[ht]
    \setlength{\tabcolsep}{27.7pt}
    \renewcommand{\arraystretch}{1.0}
    \centering
    \caption{Results of ablation studies on AgeDB-Natural.
        The best results for each metric are highlighted with \textbf{bold font}.}
    \begin{tabular}{lcccc}
        \toprule   
        Methods                              & MAE$ \downarrow  $ & MSE $\downarrow$ & G-means $\downarrow$ & $ \mathrm{R}^2 $ $\uparrow$ \\
        \midrule   
        Vanilla                              & 7.044              & 85.505           & 4.449                & 0.715                       \\
        \midrule   
        $\gamma=1, \text{proj. dim.}=64$     & 6.767              & 78.99            & 4.264                & 0.737                       \\
        $\gamma=1, \text{proj. dim.}=128$    & 6.924              & 81.61            & 4.328                & 0.728                       \\
        $\gamma=1, \text{proj. dim.}=256$    & 6.779              & 79.38            & 4.289                & 0.735                       \\
        $\gamma=1, \text{proj. dim.}=512$    & \textbf{6.724}     & \textbf{77.53}   & \textbf{4.245}       & \textbf{0.741}              \\
        $\gamma=1, \text{proj. dim.}=1024$   & 6.824              & 80.06            & 4.283                & 0.733                       \\

        \midrule
        $\text{proj. dim.}=512, \gamma=0.01$ & 6.764              & 77.65            & 4.215                & 0.741                       \\
        $\text{proj. dim.}=512, \gamma=0.1$  & 6.866              & 80.76            & 4.368                & 0.731                       \\
        $\text{proj. dim.}=512, \gamma=1.0$  & \textbf{6.724}     & \textbf{77.529}  & \textbf{4.245}       & \textbf{0.741}              \\
        $\text{proj. dim.}=512, \gamma=10$   & 6.879              & 81.146           & 4.331                & 0.729                       \\
        $\text{proj. dim.}=512, \gamma=100$  & 6.859              & 80.819           & 4.276                & 0.731                       \\
        \bottomrule   
    \end{tabular}
    \label{tabS1}
\end{table*}

\subsection{B5. Baseline Implementation Details}

For IMDB-WIKI and AgeDB experiments, the vanilla model's training configuration aligned closely with ACCon, using the Adam optimizer, a batch size of 64, an initial learning rate of 0.00025, and the same decay scheme.
NaïveSupCon adopted hyperparameters consistent with ACCon.
AdaSupCon set the weight coefficient $\gamma$ to 0.75, as per reported optimal performance \cite{dai2021adaptive}.
RankSIM used  $\gamma=100$ and interpolation strength of 2.
BMSE \cite{ren2022balanced}, which uses a batch-based Monte Carlo implementation as the loss function, initialized the noise parameter $\sigma$ to 1.
LDS applied a Gaussian smooth kernel ($\text{size} = 5$, $\sigma = 2$).
RNC \cite{zha2024rank} was first trained using self-supervised learning with similar data augmentation as our approach, then fine-tuned using L1 loss.
ConR parameters were set to similarity $\text{window} = 1$ and $\beta=3$, consistent with \cite{keramati2023conr}.

For STS-B-Natural, the vanilla model used a batch size of 128, learning rate of 0.0001 with Adam optimizer, validating every 400 iterations, with training stopping if the validation error did not decrease after 30 consecutive checks.
STS-B-DIR results were referenced from \cite{gong2022ranksim}. AdaSupCon and NaïveSupCon used same settings to ACCon for both STS-B-Natural and STS-B-DIR.
RankSIM was trained with a batch size of 16, learning rate of 0.00025, $\gamma=0.0003$, and interpolation = 2.
Training would stop if the validation error did not decrease after 30 consecutive checks, with a maximum of 300 checks allowed.
RankSIM results on STS-B-DIR were quoted from \cite{gong2022ranksim}.
Re-weighting methods (LDS, Focal-R) followed the setup in \cite{yang2021delving}.

\newpage

\section{Appendix C. Supplementary Results}
\subsection{C1. Ablation Studies}

The results of the supplemental ablation studies are shown in the Table \ref{tab55}.

\begin{table}[ht]
    \setlength{\tabcolsep}{7.6pt}
    \renewcommand{\arraystretch}{1.0}
    \caption{Results of ablation experiments on the AgeDB-Natural.
        We use the \textbf{bold} font to illustrate the best results for each metric.}
    \centering
    \small
    \begin{tabular}{lcccc}
        \toprule   
        Methods      & MAE $\downarrow$\quad & MSE $\downarrow$\qquad & G-means $\downarrow$\qquad & $ \mathrm{R}^{2} $  $\uparrow$\qquad \\
        \midrule   
        Before proj. & 6.855                 & 81.31                  & 4.367                      & 0.729                                \\
        W/o proj.    & 6.982                 & 83.79                  & 4.434                      & 0.721                                \\
        Two-stage    & 9.499                 & 146.0                  & 6.213                      & 0.513                                \\
        ACCon (Ours) & \textbf{6.724}        & \textbf{77.53}         & \textbf{4.245}             & \textbf{0.741}                       \\
        \bottomrule   
    \end{tabular}
    \label{tab55}
\end{table}

\subsection{C2. Visualization of Ablation Studies}
\label{vis_abla}

Figure \ref{figS1} presents t-SNE visualizations of feature representations under various settings on the AgeDB-Natural test set.
Notably, using features before the projection layer yields tight, continuous features but decreases correlations between $\cos\left(\theta_{i,j}\right)$ and $\left|y_{i}-y_{j}\right|/100$.
Conversely, removing the projection layer enhances representation and label distance relationships but reduces cohesiveness.
Compared to ablation settings, our framework achieves a balance, maintaining tight, continuous features while preserving correlations  between $\cos\left(\theta_{i,j}\right)$ and $\left|y_{i}-y_{j}\right|/100$.

\subsection{C3. Hyper-parameters Sensitivity Analysis}
\label{hyper-parameters}

To evaluate the sensitivity of our method to hyper-parameters, we evaluated sensitivity to the weight coefficient ($\gamma$) and projection dimension.
We conducted performance evaluations on AgeDB-Natural using $\gamma$ values of 0.01, 0.1, 1, 10, and 100, and projection dimensions of 64, 128, 256, 512, and 1024.
Table \ref{tabS1} shows our method consistently outperformed the vanilla model across all hyper-parameter combinations, demonstrating robustness to hyper-parameter variations.

\subsection{C4. Reproducibility of Main Results}
\label{Reproducibility}
To account for random seed influences, we repeated experiments on AgeDB-Natural 3 times.
Table \ref{tabS2} shows results consistent with those reported in the main paper.

\begin{table}[t]
    \caption{The average metrics and its standard deviations on AgeDB-Natural.}
    \centering
    \small
    \begin{tabular}{cccc}
        \toprule   
        MAE $\downarrow$\quad & MSE $\downarrow$\qquad & G-means $\downarrow  $\qquad & $ \mathrm{R}^{2}$ $\uparrow$\qquad \\
        \midrule   
        6.724 $\pm$ 0.08      & 77.53 $\pm$ 2.72       & 4.245 $\pm $ 0.06            & 0.741 $\pm$ 0.01                   \\
        \bottomrule   
    \end{tabular}
    \label{tabS2}
\end{table}

\section{Appendix D. Limitations}
\label{Limitations}

Our proposed method has two primary limitations.
\begin{itemize}
    \item Its application to dense pixelwise regression tasks is constrained by high computational resource requirements, particularly in tasks involving precise depth estimation of images.
    \item Current methods require labeled training data, which may not be universally available.
\end{itemize}

These limitations are acknowledged, and future work will address these issues to expand the method's applicability.

\newpage

\bibliography{refs}